\documentclass[journal]{IEEEtran}
\usepackage{color}
\usepackage{bbding}
\usepackage{pifont}
\usepackage{amssymb}
\usepackage{amsthm}
\usepackage{booktabs}
\usepackage{multirow} 
\usepackage{setspace}
\usepackage{epsfig}
\usepackage{graphicx}
\usepackage{subfigure}
\usepackage{algorithm}
\usepackage{algpseudocode}
\usepackage{url}
\usepackage{amsmath}
\usepackage{bm}
\usepackage{tablefootnote}
\usepackage{float}

\usepackage{array}
\newcolumntype{L}[1]{>{\raggedright\let\newline\\\arraybackslash\hspace{0pt}}m{#1}}
\newcolumntype{C}[1]{>{\centering\let\newline\\\arraybackslash\hspace{0pt}}m{#1}}
\newcolumntype{R}[1]{>{\raggedleft\let\newline\\\arraybackslash\hspace{0pt}}m{#1}}

\usepackage{xcolor}

\ifCLASSINFOpdf
\else
\fi
\begin{document}
\thispagestyle{empty}


	%
	\title{Keeping Deep Lithography Simulators Updated: Global-Local Shape-Based Novelty Detection and Active Learning}
	%
	%
	%
	
	\author{Hao-Chiang~Shao,~\IEEEmembership{Member,~IEEE}, Hsing-Lei Ping, Kuo-shiuan Chen, Weng-Tai Su,~\IEEEmembership{Member,~IEEE}, Chia-Wen~Lin,~\IEEEmembership{Fellow,~IEEE}, Shao-Yun~Fang,~\IEEEmembership{Member,~IEEE}, 
	Pin-Yian~Tsai, and~Yan-Hsiu~Liu
	\thanks{Manuscript received on January 22, 2022. (Corresponding Author: Chia-Wen Lin)}
	\thanks{Hao-Chiang Shao is with the Department of Statistics and Information Science, Fu Jen Catholic University, Taiwan. (e-mail:shao.haochiang@gmail.com)}
	\thanks{Hsing-Lei Ping, Kuo-shiuan Chen and Weng-Tai Su were with the Department of Electrical Engineering, National Tsing Hua University, Hsinchu, Taiwan.}		
	\thanks{Chia-Wen Lin is with the Department of Electrical Engineering and the Institute of Communications Engineering, National Tsing Hua University, Hsinchu, Taiwan. (e-mail: cwlin@ee.nthu.edu.tw)}
	\thanks{Shao-Yun Fang is with the Department of Electrical Engineering, National Taiwan University of Science and Technology, Taipei, Taiwan. (e-mail: syfang@mail.ntust.edu.tw)}
	\thanks{Pin-Yian Tsai and Yan-Hsiu Liu are with United Microelectronics Corporation, Hsinchu, Taiwan. (e-mail: \{pin\_yian\_tsai; cecil\_liu\}@umc.com)}
	\thanks{Color versions of one or more of the figures in this paper are available online at http://ieeexplore.ieee.org.}
	}
	
	%
	%

	\markboth{IEEE Transactions on Computer-Aided Design of Integrated Circuits and Systems,~Vol.~x, No.~x, Month~2022}%
	{Shell \MakeLowercase{\textit{et al.}}: Bare Demo of IEEEtran.cls for IEEE Journals}
	%



	\maketitle
	
	\begin{abstract}
	
    Learning-based pre-simulation (i.e., layout-to-fabrication) models have been proposed to predict the fabrication-induced shape deformation from an IC layout to its fabricated circuit. Such models are usually driven by pairwise learning, involving a training set of layout patterns and their reference shape images after fabrication. 
However, it is expensive and time-consuming to collect the reference shape images of all layout clips for model training and updating.
To address the problem, we propose a deep learning-based layout novelty detection scheme to identify novel (unseen) layout patterns, which cannot be well predicted by a pre-trained pre-simulation  model.
We devise a global-local novelty scoring mechanism to assess the potential novelty of a layout by exploiting two subnetworks: an autoencoder and a pretrained pre-simulation model. The former characterizes the global structural dissimilarity between a given layout and training samples, whereas the latter extracts a latent code representing the fabrication-induced local deformation. 
By integrating the global dissimilarity with the local deformation boosted by a self-attention mechanism, our model can accurately detect novelties without the ground-truth circuit shapes of test samples.
Based on the detected novelties, we further propose two active-learning strategies to  sample a reduced amount of representative layouts most worthy to be fabricated for acquiring their ground-truth circuit shapes.
Experimental results demonstrate i) our method's effectiveness in layout novelty detection, and ii) our active-learning strategies' ability in selecting representative novel layouts for keeping a learning-based pre-simulation model updated.
	
	\end{abstract}
	

	%
	\IEEEpeerreviewmaketitle

	\section{Introduction}
	\label{sec:intro}

After integrated circuit (IC) circuit design and layout, it needs a multi-step sequence of photolithographic and chemical processing steps to  fabricate an IC wafer. Because of the great exposure variations in the lithography procedure and the chemical reactions in the etching procedure during fabrication, the lithography and etching procedures together result in nonlinear shape deformation of a designed IC pattern, which is usually too complicated to model. This fact therefore urges the development of deep learning-based pre-simulation models, such as GAN-OPC~\cite{yang2019gan}, LithoGAN~\cite{ye2019lithogan}, and LithoNet-OPCNet~\cite{shao2021ic}, to handle issues like i) lithography simulation for predicting the shapes of fabricated circuit based on a given IC layout along with IC fabrication parameters, and ii) mask optimization for predicting the best mask to compensate for the fabrication-induced shape deformations. 
However, deep learning-based models and their training/updating processes usually rely on tremendous training data. The selection of training data consequently becomes a considerable issue because whether a training dataset or a fine-tune dataset (with novel patterns) is informative largely affects the generalization ability of a learning-based model.

\begin{figure*}[!t]
    \centering
    \includegraphics[width=0.85\textwidth]{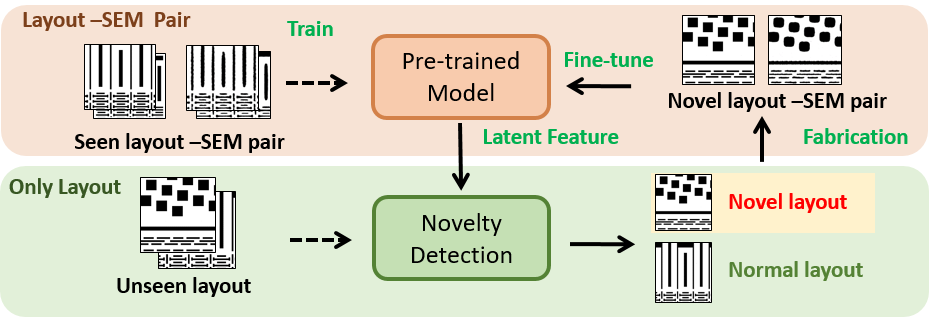}
    \caption{Relationship  among  novelty detection,  pretrained simulation model, newly-designed layout, and IC fabrication.  Given a pretrained pre-simulation model, e.g. LithoNet, driven by layout---SEM image pairs, the proposed layout novelty detection method aims to identify potential novel layouts from a pool of newly-designed layout patterns so that these potential novel layouts can be further fabricated to derive their SEM images for fine-tuning the pretrained model. Because only selected layouts need to be fabricated, this framework can save budgets and times for collecting good enough training samples. The proposed novelty detection method can thus act as an oracle for active learning.  
    }
    \label{fig:concept}
\end{figure*}

However, how to collect an appropriate training dataset and sample a fine-tune dataset, also known as a development set, from the IC layout design and fabrication processes is practically a very complicated issue. This complexity is due to the following two aspects. 
First, given a deep learning model pretrained on an initial training dataset, it may still need to be fine-tuned on another development set so that it can be generalized to  those samples that are unseen in the initial training dataset.
In order to determine a proper development dataset, one has to assess a layout's degree of novelty by checking whether this layout's SEM (Scanning Electron Microscope) image shape can be accurately predicted by the pretrained pre-simulation model even in the absence of this layout's ground-truth SEM image. 
Second, it is unaffordable to exhaustively collect the aerial images of each layout under different fabrication parameter settings as training ground-truths because of high costs. 
For example, LithoNet~\cite{shao2021ic} learns the layout-to-SEM contour correspondence and the effects of fabrication parameters from a collection of layout-SEM image pairs. If LithoNet needs to learn how $1,000$ layout patterns deforms under $7$ different fabrication parameter settings, one should fabricate $1,000 \times 7$ IC circuits to obtain a comprehensive training set covering all $7,000$  combinations of to-be-learned conditions. Such fabrication processes required to collect training data are too time-consuming and costly.
Consequently, these two aspects bring the demand for novelty detection and  active learning to both the maintenance of a learning-based pre-simulation model and the selection of a development set, as illustrated in Fig.~\ref{fig:concept}.

To address this issue, we propose an SEM-free (i.e., ground-truth-free) scheme to detect novel layout patterns, whose SEM images are worthy of collecting via the costly IC fabrication process, for informatively updating a pretrained DNN-based lithography simulator. 
That is, given i) a deep-learning-based pre-simulation model, \textit{e.g.}, LithoNet~\cite{shao2021ic}, and ii) a pool of newly-designed IC layout clips, the proposed method aims to identify layout patterns, which are novel and informative to increase the generalization ability of the pretrained model. 
This scenario leads to two considerations. 
First, the method should be able to identify novel (unseen) layout designs whose fabrication-induced shape deformation cannot be well predicted by the model pretrained on an initial training set of layout-SEM image pairs. The difficulty laying behind the first consideration is that, during deployment, all inputs are layout patterns, and therefore our method needs to detect layout novelty by learning the relationship between layouts and their predicted layout-to-SEM deformation maps. 
Second, due to the high costs of IC fabrication and taking SEM images, the method should be able to select a reduced set of most informative layouts to update the pretrained model for the sake of budget efficiency. As a result, only the selected set of layouts will be fabricated to acquire their layout-SEM pairs for fine-tuning the pretrained model. This consideration also hints at the requirement of a sampling process for active learning. Because the solution of the second consideration, i.e., the capability of selecting data samples that can optimally represent the entire training data domain, highly depends on that of the first, we 
take into account both considerations to propose novel methods for novelty detection and active learning. 
This work has the following major contributions. 


\noindent $\bullet$ Our novelty detection method is the first to learn global-local features for identifying which layouts are novel and worthy of further fabrication, in contrast to existing approaches that detect novelties based on solely global features \cite{pimentel2014review,salehi2021unified}. Specifically, we devise two subnetworks to derive two novelty scores---one for measuring global structure dissimilarity and the other for capturing local deformation---complementary to each other. In this way, our method can efficiently collect informative layout--SEM image pairs, which are necessary for fine-tuning learning-based layout-to-SEM prediction or mask-optimization models like LithoNet/OPCNet to keep them updated with newly designed data. \\
\noindent $\bullet$ During deployment, our method can detect novel layout  patterns in the absence of the ground-truth SEM images of target layouts' fabricated circuits. Therefore, our model not only meets the practical field requirements for layout pre-inspection, but also functions as an active learning oracle.\\ 
\noindent $\bullet$ We further propose two effective graph sampling-based active-learning strategies, namely one-time sampling and incremental sampling, to sample a much reduced set of representative layouts, which are most worthy of further fabrication for acquiring their reference SEM images, in an on-a-budget environment.



The  remainder  of  this  paper  is  organized  as  follows.  We review related literature in Sec.~ \ref{sec:review}. The proposed layout novelty detection method is detailed in Sec. \ref{sec:method}. Sec.~\ref{sec:sampling} presents our proposed active learning strategies. Sec.~\ref{sec:exp} demonstrates and discusses our experimental results. Finally, we draw our conclusion in Sec. \ref{sec:conclusion}.
	
	\section{Related Work}
	\label{sec:review}
	\subsection{Learning-Based Lithography Pre-simulation Models}

Several learning-based lithography pre-simulation models were proposed for topics such as lithography simulation and mask optimization. 
In order to save the computational resources, Yang \textit{et al.} proposed the GAN-OPC method \cite{yang2019gan} to facilitate the mask optimization process. GAN-OPC aims at creating quasi-optimal masks for given target circuit patterns by learning target-mask mappings. GAN-OPC can  generate high-quality masks and thus ensure good printability while requiring reduced normal OPC steps. In addition, Ye \textit{et al.} devised LithoGAN for lithography simulation \cite{ye2019lithogan}. LithoGAN is a GAN-based end-to-end lithography modeling framework that maps input mask patterns directly to the output resist patterns, making it capable of predicting resist patterns accurately while achieving significant speedup compared with conventional lithography simulation methods. Recently, our proposed LithoNet-OPCNet framework~\cite{shao2021ic}, successfully  addresses the lithography simulation and mask optimization problems simultaneously in an end-to-end learning manner. 
Specifically, LithoNet, trained on a comprehensive set of layout--SEM image pairs, can accurately predict for an input layout pattern the fabrication-induced shape distortion, 
OPCNet~\cite{shao2021ic}, trained with the guidance provided by a pretrained LithoNet, aims to predict the optical proximity corrected (OPC) photo-mask pattern of an input layout.





\subsection{Novelty Detection}

Novelty detection is the procedure used to identify if a data sample is hitherto unknown. 
It is typically modeled as a one-class classification problem, in which a novelty detector is trained on single-class training samples which are all supposed to be seen normal ones. As a result,  the detector can determine whether an input testing sample is dissimilar to the seen training samples in terms of a given distance metric \cite{liu2013svdd,terrell1992variable}  or a loss function \cite{japkowicz1995novelty} during deployment. 
Though novelty detection is closely related to anomaly/outlier detection, their scenarios are significantly different. Specifically,  anomaly/outlier detection methods usually learn to find abnormal samples in a given reference dataset, whereas the reference data used to train a novelty detection model are assumed to be \textit{unpolluted} and involve only normal regular samples.  
Note that a novelty detection method, like other anomaly/outlier detection methods, usually maps its input data to a novelty score so that an appropriate threshold can be defined accordingly to tell novel samples (outliers) and regular ones (inliers) apart~\cite{miljkovic2010review}.



Novelty detection methods have found applications in video surveillance~\cite{calderara2011detecting,mathieu2015masked}, medical imaging~\cite{schlegl2017unsupervised}, abnormal event detection for attributed network~\cite{fan2020anomalydae,wang2020one}, etc. 
In general, common approaches based on 
probabilistic-based models, such as one-class SVMs~\cite{liu2013svdd} and kernel density estimation~\cite{terrell1992variable}, can achieve good performances on handling low-dimensional features. However, these methods may not apply to high-dimensional data well, \textit{e.g.}, images in computer vision tasks. 
Hence, two sorts of CNN (convolutional neural network) based methods have been proposed to address this problem. One sort learns to generate a reconstructed image and then evaluate an abnormal score according to the difference between the input and the reconstructed images~\cite{an2015variational,kliger2018novelty}, and the other  learns to embed a latent structural feature of the input and then derive an abnormal score based on the extracted structural feature~\cite{abati2019latent}. 

For example, Sabokrou \textit{et al}. proposed to train an auto-encoder along with a discriminator, conceptually a classifier, in an adversarial manner based on the reconstruction error, and then to determine whether an input is novel by the discriminator \cite{sabokrou2018adversarially}.
Similarly, Perera \textit{et al}.  proposed OCGAN \cite{perera2019ocgan} to solve the one-class novelty detection problem by learning the latent representations of within-class examples via a denoising auto-encoder network. 
Moreover,  DSGAN was proposed in \cite{sung2019difference} for synthesizing novel samples surrounding real training data such that the decision border between the regular data and novelties can be determined effectively by typical models. 
Besides, Pidhorskyi \textit{et al}. devised an architecture consisting of an auto-encoder and a discriminator for anomaly detection~\cite{pidhorskyi2018generative}. Their model is trained on top of a double min-max-game framework that iteratively optimizes the distribution of latent codes extracted by the auto-encoder and the fidelity of the reconstructed images.


Furthermore, classification-based novelty/anomaly detection models can generally be boosted via a self-supervised mechanism~\cite{golan2018deep,bergman2020classification,tack2020csi}. The applications of these methods are, however, limited by the assumption that their pre-processing strategies, \textit{e.g.}, rotation, random cropping, and geometric transformations, cannot alter the ground-truth labels (\textit{i.e.}, class information) of the training dataset. As for IC fabrication, a rotated or geometrically-transformed layout pattern will result in a different printed images  because 
the processing results of a stepper/scanner in the x- and y-directions are asymmetric, and hence the self-supervised mechanism is usually not applicable. 

\subsection{Active Learning}

Active learning refers to cases in which a learning algorithm can assess the necessity of labeling an unlabeled sample by interactively querying an oracle---usually a pre-trained model or a user-specified metric function---about unlabeled samples' importance~\cite{settles2009active}. 
A fundamental concept is uncertainty-based selection, through which an oracle recommends (unlabeled) data of high-uncertainty for labeling and disregards those high-confidence ones \cite{lewis1995sequential,tong2001support}. This sort of methods are, however, sensitive to outliers. 
Recently, several active learning algorithms were devised for CAD/VLSI applications, such as the methods in \cite{lin2018data,zhuo2010active}. 
However, all these active learning techniques need to collaborate with a reliable oracle. Hence, we aim in this paper to develop an oracle that can assess the novelty of an unseen layout in an SEM-free environment by learning the knowledge contained in pairwise training samples. 

	\section{Global-Local Shape-Based Novelty Detection}
	\label{sec:method}
\begin{figure*}[!t]
    \centering
    \includegraphics[width=0.7\textwidth]{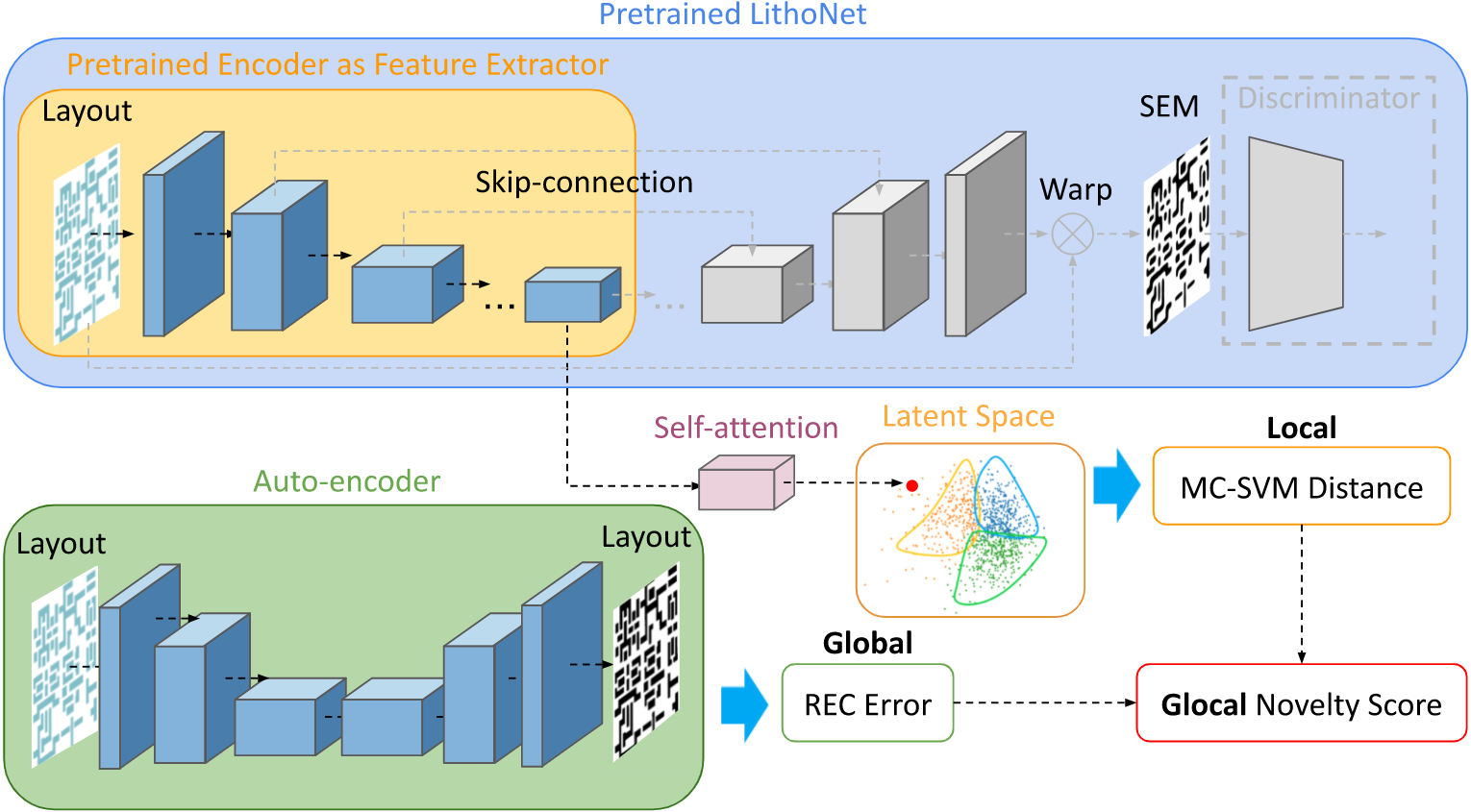}
    \caption{Framework of the proposed layout novelty detection method. Upper part: The SA-LithoNet, architecturally the encoder part of pretrained LithoNet followed by a self-attention module. Lower part: The autoencoder. The SA-LithoNet can embed an input layout into a latent code characterizing the local layout-to-SEM deformations, whereas the autoencoder is used to measure the global dissimilarity via the reconstruction error. These two parts can jointly derive a global-local (\textbf{Glocal}) score for layout novelty detection.}
    \label{fig_2}
\end{figure*}

\subsection{Overview}

Due to limited labeling resources, 
one often adopts a sampling strategy to select a small set of most informative unlabeled novel samples for further labeling routine and then use the newly labeled samples to update the learned model in an active learning manner. 
As reported in \cite{zhou2017fine}, while regular samples, \textit{e.g.}, layout clips, with characteristics similar to that of source training data can usually be predicted  fairly well by a model pre-trained on the same source training set, data with unseen patterns, \textit{e.g.}, novel layout clips, are potentially able to improve a pretrained model and thus worth a fabrication to acquire their ground-truth SEM images.
Hence, under the premise of saving the costs of fabricating excessive training samples and acquiring their SEM images, our goal here is to identify the most informative unseen layout clips, which are worth a fabrication for acquiring their SEM images to effectively update a pretrained model, from a pool of newly-designed IC layout clips. 

%
%

To tackle this active learning problem for an IC fabrication pre-simulation model like LithoNet~\cite{shao2021ic}, we aim to design a layout novelty detection scheme that can work in the absence of ground-truth SEM images during the inference stage. It can distinguish \textbf{novel} layout clips, whose SEM images cannot be accurately predicted by a pre-simulation model (\textit{e.g.},  LithoNet~\cite{shao2021ic}), from \textbf{regular} layouts whose SEM images can be well predicted. 
To this end, we elaborate first in Sec.~\ref{subsec:301} our supervised scheme to label novel layout patterns objectively by annotating novel regions on layout clips with the aid of ground-truth SEM images. We then describe our unsupervised layout novelty detection scheme, namely \textbf{Glocal novelty score}, in Sec.~\ref{subsec:glocalnovelty}--\ref{subsec:global}.

Fig.~\ref{fig_2} shows the architecture of our proposed \textbf{Glocal} (global-local) method that consists of two primary components, \textit{i.e.}, an SA-LithoNet and an autoencoder. 
Suppose that a novel layout should result from the innovation of global planning, the change of local planning, or both. Our method exploits i) LithoNet, a pre-simulation model of fabrication-induced local shape deformation~\cite{shao2021ic}, for capturing local shape features with the aid of a self-attention (SA) module, and ii) an autoencoder for characterizing  global shape properties. 
The SA-LithoNet is architecturally the encoder part of a pretrained LithoNet followed by a self-attention module. This design employs SA-LithoNet to  extract a  feature representing local layout-to-SEM deformations within attended regions, identified by the self-attention module supervised by the novelty labels. Based on the assumption that, the local-shape feature of a novel sample should be deviated from the distribution of regular samples, we employ the SA-LithoNet feature for local novelty scoring via multi-class SVM (MC-SVM) classification. Besides that,  we use the reconstruction error with the autoencoder, representing the global shape dissimilarity, as the global novelty score. As a result, we combine the local and global novelty scores to obtain the \textbf{Glocal}  novelty score. 



\subsection{Model Inconsistency-Guided Novelty Annotation}
\label{subsec:301}
Because manually annotating novel patterns in a rich collection of layout clips is nontrivial, even for an experienced engineer, we devise first a supervised mechanism for annotating potential novelties on layout patterns to train and evaluate our novelty detection model.
This mechanism aims to find a novel layout pattern based on the inconsistency between the pattern's ground-truth SEM image and the corresponding layout-to-SEM prediction yielded by LithoNet~\cite{shao2021ic}. Hence, we name this mechanism \textbf{Model Inconsistency-Guided Novelty Annotation (MIGNA)}.

MIGNA aims to identify those local regions where the shape contours of the layout-to-SEM predictions~\cite{shao2021ic} significantly deviate from their counterparts in the corresponding ground-truth SEM images. Such deviations imply that the pretrained  layout-to-SEM prediction model may not have learned from enough similar training layout patterns yet. Common sorts of shape deviations may include, for example, unexpected abnormal patterns such as enclosures, neckings, and bridges. 
One possible cause of these deviations is the unexpected diffraction, usually induced by the layout arrangements around abnormal patterns, during the lithography process, which makes the same layout pattern result in different SEM patterns with neighborhood-related shape variants. Consequently, when a learning-based pre-simulation model like LithoNet is trained on a training dataset containing insufficient similar patterns, its shape predictions tend to deviate from the corresponding ground-truths. Such deviations should be considered as anomalies, i.e., layout novelties, due to insufficient training patterns.

To identify unexpected shape deformations due to inaccurate predictions, we set a threshold of three standard deviations from the mean L1-norm of the pixel-wise differences between the layout-to-SEM predictions and their ground-truth SEM images. Three standard deviations from the mean is a common cut-off in practice for identifying outliers in a Gaussian-like distribution\footnote{Statistically, $99.7\%$ data fall within $\mu \pm 3\sigma$, and thus the rest $0.3\%$ data are usually regarded as outliers.}. Consequently, our MIGNA method involves
the following steps. \\
\noindent \textbf{Step-1}:  Measure the  pixel-wise deformation map based on the L1-distance between the ground-truth SEM image and the layout-to-SEM prediction for the same layout clip, where  LithoNet~\cite{shao2021ic} is adopted as the layout-to-SEM predictor. \\ 
\noindent \textbf{Step-2}:  Partition the deformation map into $16^2=256$ non-overlapping patches, and discard those patches reaching image borders (60 border patches are omitted in our implementation). 
\\
\noindent \textbf{Step-3}:  Annotate a patch as ``anomaly'' if its local L1-distance exceeds the mean L1-distance of the whole training dataset by three standard deviations or more. 
\\
\noindent \textbf{Step-4}: Label a layout as ``novelty'' if it contains at least a predetermined number of abnormal patches.

In this way, we can annotate the layout novelties systematically in a supervised manner.

\subsection{Global-Local (Glocal) Novelty Score}
\label{subsec:glocalnovelty}

Inspired by residue-based and classification-based novelty detection models, as illustrated in Fig.~\ref{fig_2}, our method consists of two subnetworks: i) an autoencoder, trained on a collection of layout images, and ii) an attention-guided layout-to-SEM prediction model, SA-LithoNet, comprising the encoder part of LithoNet~\cite{shao2021ic}
and a self-attention (SA) module.  While the autoencoder characterizes the global shape appearance of a given layout, the SA-LithoNet extracts a latent feature code representing local shape deformations.
%
Then, we evaluate the \textit{global-local (Glocal)} novelty score based on i) a local anomaly score $\Theta_{\mathrm{local}}$ obtained by using SA-LithoNet (elaborated in Sec. \ref{subsec:attention} and Sec. \ref{subsec:local}), and ii) a global novelty score $\Theta_{\mathrm{global}}$ derived by using the autoencoder (elaborated in Sec. \ref{subsec:global}). 
The local anomaly score $\Theta_{\mathrm{local}}$ is derived by the proposed MC-SVM (Multi-class SVM) algorithm that estimates the distance from the training dataset to the input in the latent feature space. Meanwhile, the global novelty score $\Theta_{\mathrm{global}}$ is evaluated based on conventional residue-based novelty detection scheme.

The \textit{Glocal} novelty score 
 of an input layout $\mathbf{y}$ is defined as 
\begin{equation}
    \Theta_{\mathrm{novel}}(\mathbf{y}) =
    \mathrm{norm}(\Theta_{\mathrm{local}}(\mathbf{y})) + 
    \mathrm{norm}(\Theta_{\mathrm{global}}(\mathbf{y})) 
    \mbox{,}
    \label{eq:novelscore}
\end{equation}
where $\mathrm{norm}(\cdot)$ denotes the following normalization process: 
\begin{equation}
    \mathrm{norm}(\Theta(\mathbf{y})) = \frac{\Theta(\mathbf{y}) - {\mathrm{mean}}(\Theta(\mathbf{y}))}{\mathrm{std}(\Theta(\mathbf{y}))} \mbox{,}
    \label{eq:normalization}
\end{equation}
where $\mathrm{mean}(\cdot)$ and $\mathrm{std}(\cdot)$ denote the mean and standard deviation, respectively. 
Note that (\ref{eq:novelscore}) follows the designs in \cite{abati2019latent,pidhorskyi2018generative}, in which a final novelty score is obtained by summing up two normalized independent novelty scores.


\subsection{Attention-Guided Layout-to-SEM Prediction Model}
\label{subsec:attention}

\begin{figure}[!t]
    \centering
    \includegraphics[width=0.48\textwidth]{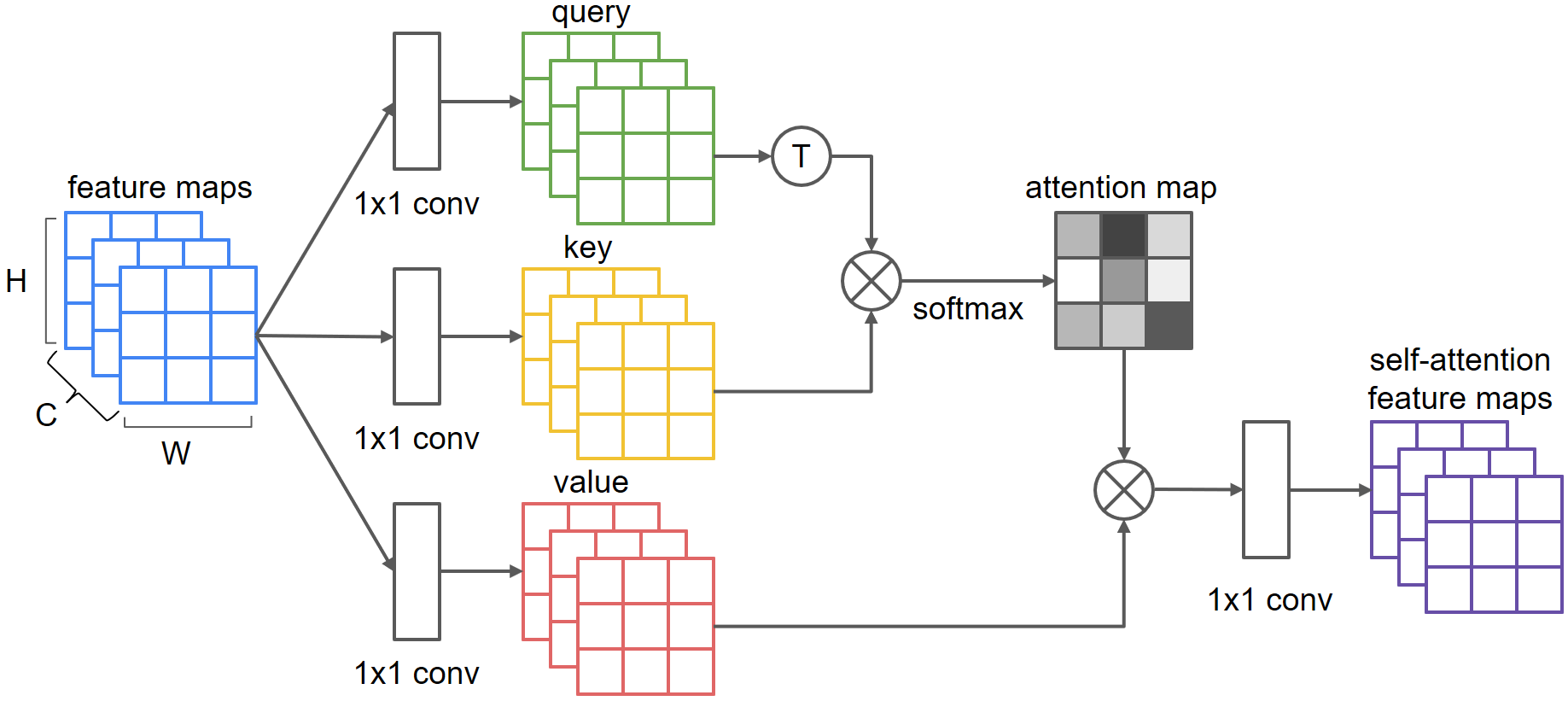}
    \caption{Block diagram of self-attention module for learning the dependencies between the novelty label and LithoNet features in our proposed attention-guided novelty detection model.}
    \label{fig_self_attention}
\end{figure}

Self-attention (SA) mechanisms~\cite{vaswani2017attention} like Vision Transformer~\cite{dosovitskiy2020image} and Non-local Neural Networks~\cite{wang2018nonlocal} have recently demonstrated their high efficacy in finding spatial long-range dependencies among image patches so that all dependent contextual features can be taken into account together to optimize a specific vision task. In order to extract a latent code carrying wider-range representative features for characterizing the fabrication-induced circuit shape deformation, we propose SA-LithoNet by appending an SA module to the encoder of LithoNet~\cite{shao2021ic}, as illustrated in Fig.~\ref{fig_self_attention}.

By evaluating the dependencies between patches within the latent feature tensor embedded by the encoder of LithoNet, the SA module reorganizes the latent feature and then takes into account a wider-range of layout shape details according to the patch dependencies, as will be described later in (\ref{eq:SA_beta}). 
Besides, to reduce the amount of parameters while still achieving a good performance, we adopt the design of SAGAN~\cite{zhang2019self} and replace the fully connected layer with $1\times 1$ convolutions, based on which the \textit{query} ($\mathbf{Q}$), \textit{key} ($\mathbf{K}$), and \textit{value} ($\mathbf{V}$) maps are derived from dimension-reduced features. 
As illustrated in Fig.~\ref{fig_self_attention}, the self-attention module can be expressed as 

\begin{equation}
    \left[
    \begin{array}{ccc}
         \mathbf{Q} & \mathbf{K} & \mathbf{V}\\
    \end{array} 
    \right] 
    = \left[
    \begin{array}{ccc}
         \mathbf{W}^T_q & \mathbf{W}^T_k & \mathbf{W}^T_v\\
    \end{array}
    \right] \mathbf{f}_\mathrm{Litho}(\mathbf{x}) \mbox{,}
    \label{eq:selfAtt_qkv}
\end{equation}
where $\mathbf{f}_\mathrm{Litho}(\mathbf{x})$ denotes the $C\times (H\cdot W)$ latent feature extracted from  input layout pattern $\mathbf{x}$ by LithoNet~\cite{shao2021ic}, $H$ and $W$ are respectively the height and width of the feature and $C$ is the  feature channel-depth, $C^{'}=\frac{C}{8}$, $W_{v}\in\mathbb{R}^{C\times C}$, $W_{q}\in\mathbb{R}^{C \times C^{'}}$, and $W_{k}\in \mathbb{R}^{C \times C^{'}}$ are $1\times1$ convolution kernels for feature channel-depth reduction.

As shown in Fig.~\ref{fig_self_attention}, the attention map derived after softmax is
\begin{equation}
    \beta_{j,i} = \frac{e^{s_{ij}}}{\sum_{i=1}^{HW}\sum_{j=1}^{HW} e^{s_{ij}}}\mbox{.}
    \label{eq:SA_beta}
\end{equation}
where, $s_{ij} = \mathbf{q}_{i}^{T}\mathbf{k}_{j}$, $\mathbf{q}_{i}$ and $\mathbf{k}_{j}$ are $C\times 1\times 1$ tensors, $\beta_{j,i}$ represents the normalized attention (dependency) in the $j$-th \textit{query} tensor contributed by the $i$-th \textit{key} tensor.  
Therefore, the output self-attention feature map $\mathbf{O}=(\mathbf{o}_{1}, \mathbf{o}_{2}, \dots, \mathbf{o}_j, \dots,\mathbf{o}_{HW})$ is a $C \times (W \cdot H)$ tensor, with  $\mathbf{o}_{j}$ obtained by 
\begin{equation}
    \mathbf{o}_{j} = \mathbf{W}^T_o \, \sum_{i=1}^{HW} \beta_{j,i} \mathbf{v}_{i} \mbox{.}
\end{equation}
where $\mathbf{v}_{i}$ is the $i$-th $C\times 1 \times 1$ sub-tensor within the \textit{value} map $\mathbf{V} \in \mathbb{R}^{C\times(W\dot H)}$, and $\mathbf{W}^T_o \in \mathbb{R}^{C\times C}$ denotes the $1 \times 1$ convolution kernels. 

As a result, the final feature tensor $\mathbf{f}_\mathrm{SA}(\mathbf{x})$ enhanced by this SA module is 
\begin{equation}
    \mathbf{f}_\mathrm{SA}(\mathbf{x}) = \gamma \mathbf{O} + \mathbf{f}_\mathrm{Litho}(\mathbf{x}) \mbox{,}
    \label{eq:featureSA}
\end{equation}
where $\gamma$ is a learnable parameter, initialized as $1$. 

The SA module can learn the spatial dependency within the input feature tensor and is then used to derive a tensor more representative than its input for the novelty detection task. Based on the assumption that the local-shape feature of a novel sample should be deviated from the distribution of regular samples, we employ the SA-LithoNet feature $\mathbf{f}_\mathrm{SA}(\mathbf{x})$ in (\ref{eq:featureSA}) to evaluate the local novelty score based on the proposed Multi-Class SVM (MC-SVM) method described below.

\subsection{Local Shape Deformation-Based Novelty Score}
\label{subsec:local}

Generally, MC-SVM performs $K$-means clustering to group the training data into $K$ feature clusters at first, and then applies  one-class SVMs (OC-SVMs)  \cite{liu2013svdd,li2003improving} on the $K$ feature clusters individually to map regular-sample features into $K$ independent hyperspheres. 
Given a layout sample $\mathbf{y}$, we apply MC-SVM to evaluate its novelty score based on the distance between the sample feature $\mathbf{z}=\mathbf{f}_\mathrm{SA}(\mathbf{y})$ and each hypersphere center $\mathbf{c}_k$, where $\mathbf{f}_\mathrm{SA}(\cdot)$ denotes the attention-guided feature embedding formulated in (\ref{eq:featureSA}). If the minimal unseen-to-center distance exceeds a threshold, 
then the unseen sample $\mathbf{y}$ is classified as a novelty. 

First, in the  $K$-means clustering step of our MC-SVM-based novelty detection, given a training dataset $\mathcal{D} =\{\mathbf{x}_{1},\mathbf{x}_{2},...,\mathbf{x}_{N}\}$ and a set of latent features $\mathcal{Z} = \{ \mathbf{z} \, \mid \,\mathbf{z} = \mathbf{f}_\mathrm{SA}(\mathbf{x})\mbox{,} \,\forall \mathbf{x} \in \mathcal{D}\}$, we iteratively group all $\mathbf{z}\in\mathcal{Z}$ into $K$ clusters in the feature space and find the $K$ cluster centers by solving the following optimization problem: 
\begin{equation}
\mathop{\arg\min}_\mathbf{S} \sum_{k=1}^{K} \sum_{\mathbf{z} \in \mathcal{S}_k } \| \mathbf{z} - \boldsymbol{\mu}_k \|^2 
\mbox{,} 
\end{equation}
where $\mathbf{S} = \{\mathcal{S}_1, \mathcal{S}_2, \dots, \mathcal{S}_K\}$ with $\mathcal{S}_k$ denoting the $k$-th cluster, and $\boldsymbol{\mu}_k$ is the cluster center of $\mathcal{S}_k$.

Then, the second step of MC-SVM is to map the $K$ clusters $\mathcal{S}_k$ into $K$ individual hyperspheres. In this way, the novelty of a test layout pattern can be verified by checking if its mapped feature is far away from all $K$ hyperspheres. 
This hypersphere mapping is similar to the SVDD \cite{liu2013svdd} and OC-SVM \cite{li2003improving} algorithms. 
Specifically, for $\mathcal{S}_k$, all latent features $\mathbf{z}_{k,i} \in \mathcal{S}_k$ are mapped to a hypersphere centered at $\mathbf{c}_k$ by solving the following problem:
\begin{eqnarray}
    &&\min R_k^2 + \frac{1}{\nu}\cdot\frac{1}{m_k}\sum^{m_k}_{i=1}\xi_{k,i}  \nonumber\\
    & \mbox{subject to} & \|\phi(\mathbf{z}_{k,i})-\textbf{c}_{k} \|^2 \le R_k^2 + \xi_{k,i}
\end{eqnarray}
where $m_k=\lvert S_k \rvert$ denotes the number of samples in $\mathcal{S}_k$, 
$\xi_{k,i}$ is a slack variable used as a penalty to control the soft-boundary and the hypersphere volume with an outlier tolerance value $\nu \in (0, 1]$, $\phi$ denotes the kernel function for mapping, and $R_k$ is the radius of the $k$-th hypersphere. Numerical methods for solving this optimization problem can be found in \cite{liu2013svdd,chang2013revisit}.

As a result, we can define the local novelty score of a newly-designed layout 
 $\mathbf{y}$ as the minimal distance from its mapped latent code to the nearest hypersphere: 
\begin{equation}
    \Theta_\mathrm{local}(\mathbf{y}) = \min_k( \|\pmb{\phi}( \mathbf{f}_\mathrm{SA}(\mathbf{y}))) - \mathbf{c}_k \|^2 - R_k^{2}) \mbox{.}
\end{equation}

This local novelty score is evaluated based on the SA-LithoNet  latent code. Because LithoNet is a layout-to-SEM pre-simulation model that learns to represent local circuit shape deformations due to a fabrication process, a large $\Theta^{\mathrm{local}}$ implies that a layout sample's SA-LithoNet latent code  tends to be out-of-distribution, and that the pattern may not be predicted well with the current SA-LithoNet model. $\Theta^{\mathrm{local}}$ can thus well serve the purpose of local layout novelty scoring.


\subsection{Autoencoder-based Global Novelty Score}
\label{subsec:global}
Since the SA-LithoNet latent code is mainly for representing fabrication-induced local shape deformations, to better capture novel layout patterns, we propose to add another complimentary global feature, extracted by an autoencoder, to characterize layout patterns' global shape structures.

%
Typically, supervised by the MSE (mean-squared-error) reconstruction loss, an autoencoder learns to embed its input into a lower-dimensional latent code, based on which the autoencoder can reconstruct an image close to its input. Therefore, with the aid of the MSE loss, an autoencoder can capture the global structural characteristics of an image well. The reconstruction error between a newly-designed layout and its reconstructed version yielded by an autoencoder trained on a training dataset can thus be used to define a novelty score indicating the degree of global structural dissimilarity between the input layout and the training dataset. 
As a result, this global novelty score is defined as 
\begin{equation}
\Theta_\mathrm{global}(\mathbf{y}) = \| \mathbf{y} - \hat{\mathbf{y}} \|^{2} \mbox{,}
\end{equation}
where $\hat{\mathbf{y}}$ is the reconstructed version of $\mathbf{y}$ yielded by the autoencoder. 

	\section{Graph Sampling for Active Learning}
	\label{sec:sampling}
	%
%

After identifying novelties in a given pool of newly-designed layouts, we can then fabricate the novel layout patterns on wafers and then collect their layout-SEM pairs to update the layout-to-SEM model (\textit{e.g.}, LithoNet). However, since both fabricating ICs and taking SEM images are costly, given a limited cost budget,  we usually can only sample a small set of most representative patterns from the detected novelties for further fabrication.
To this end, we propose two sampling strategies: the one-time sampling  and the incremental sampling. Each strategy starts from building an initial undirected $k$-NN graph composed of the novel layout designs as the graph nodes 
by employing the latent code $\mathbf{f}_\mathrm{AE}$ embedded by a pretrained autoencoder. Then, based on the node degree of the initial graph, we further construct a dense graph $\mathcal{G}_d$ and a sparse graph $\mathcal{G}_s$. Finally, we rank the priority of each node via a random-walk method, whose node visiting probability is determined based on the latent code $\mathbf{f}_\mathrm{SA}$ extracted by SA-LithoNet, to select the most representative nodes accordingly.



\subsection{One-time Sampling}

The one-time sampling (OTS) algorithm aims to select the most representative layout clips from a given set of novel layout clips in only one sampling iteration. 
It primarily consists of two phases: i) data graph construction and ii) sampling by ranking. Its pseudo code is shown in Algorithm \ref{algo:vistinggraph}.

\noindent \textbf{Step-1: Data graph construction}\\
This step first estimates the data manifold, in which layout patterns lie, by building an initial $\bm{k}$-NN graph  based on the latent code $\mathbf{f}_\mathrm{AE}$ extracted by the autoencoder.
The resulting $\bm{k}$-NN graph $\mathcal{G}_{\bm{k}\mathrm{NN}}$ is a \textit{directed} graph, where each node has a fixed \textit{out-degree} of $\bm{k}$ but a variable \textit{in-degree}, and a directed edge from $p$ to $q$ represents that $q$ is a $\bm{k}$-nearest neighbor of $p$ in terms of the feature distance between $\mathbf{f}_\mathrm{AE}(p)$ and $\mathbf{f}_\mathrm{AE}(q)$. 
As a result, in order to obtain an \textit{undirected} graph specifying the distribution of layout patterns, the adjacency matrix $\mathcal{A}$ of data graph $\mathcal{G}$ is obtained by $\mathcal{A}(p,q) = \max (\mathcal{A}_{\bm{k}\mathrm{NN}}(p,q), \mathcal{A}_{\bm{k}\mathrm{NN}}(q,p))$, where $\mathcal{A}_{\bm{k}\mathrm{NN}}$ is the adjacency matrix of the initial $\bm{k}$-NN graph.


On top of $\mathcal{G}$ that characterizes the data manifold of novel layout clips, we further separate all nodes (layouts) in $\mathcal{G}$ into two groups based on each node's degree (\textit{i.e.}, the total number of edges of a node to the others) and construct one \textit{dense graph} $\mathcal{G}_{d}$ and one \textit{sparse graph} $\mathcal{G}_s$ accordingly. 
We here set $\tau_d = \mu_d + 3 \sigma_d$ as the threshold value for node separation with $\mu_d$ and $\sigma_d$ denoting respectively the mean and standard deviation of the degrees of all nodes in $\mathcal{G}$. 
Therefore, the nodes with a degree larger than $\tau_d$ are those lying in somewhere in $\mathcal{G}$ densely with similar layouts, and these nodes are used to constitute the dense graph $\mathcal{G}_{d}$. On the contrary, those nodes in $\mathcal{G}$ with a degree smaller than $\tau_d$ are used to  constitute a sparse graph $\mathcal{G}_{s}$, where each node represents a layout clip far away from other designs in the feature space. 
Note that both $\mathcal{G}_{d}$ and $\mathcal{G}_{s}$ are undirected graphs derived from $\mathcal{G}$.

\noindent \textbf{Step-2: Sampling priority ranking}\\
Because $\mathcal{G}_{d}$ and $\mathcal{G}_{s}$ contain  layout clips belonging to two different kinds of distributions, respectively, the ways to rank the sampling priorities of nodes in each graph ought to be different. 
Therefore, we devise i) two different schemes for determining starting seeds, and ii) two different weight functions for assessing the random walk probability for $\mathcal{G}_{d}$ and $\mathcal{G}_{s}$, respectively, to trigger our random-work-based graph exploration algorithm. 
Then, after exploring a given graph thoroughly, the sampling priorities of nodes in the graph are ranked by their number of total visits. 

%


The starting seeds for $\mathcal{G}_s$ and $\mathcal{G}_d$ are determined by using the \textit{closeness centrality} and the \textit{eigen-centrality}, respectively. 
This design comes from two reasons. First, because $\mathcal{G}_s$ consists of nodes (layouts) which are far from each other in the feature space, a node with a large closeness centrality, \textit{i.e.}, a small mean distance from itself to other nodes, should be representative. Second, since the nodes with higher eigen-centrality (\textit{aka} eigenvector-centrality) values in a graph  make higher impacts to other nodes as they are connected to nodes with higher eigen-centrality values~\cite{networkintroduction}, they should be sampled in higher priorities. 
The eigen-centrality of nodes on $\mathcal{G}_d$ is defined by 
\begin{equation}
    \mathcal{A}_d \mathbf{e} = \kappa_1 \mathbf{e}, \mbox{,}
    \label{eq:eigencentral}
\end{equation}
where $\mathbf{e}$ is the eigenvector recording the eigen-centrality, and $\kappa_1$ is the largest eigenvalue of $\mathcal{A}_d$, the adjacency matrix of $\mathcal{G}_d$. Also, the closeness of node $p$ in $\mathcal{G}_s$ is evaluated by
\begin{equation}
    C_{p} = \frac{n}{\sum_{q\in \mathcal{N}(p)} \mathrm{dist}_\mathrm{geo}(p,q)} \mbox{,}
    \label{eq:closeness}
\end{equation}
where $\mathrm{dist}_\mathrm{geo}(p,q)$ is the geodesic distance, i.e., the length of the shortest path on the graph, 
between nodes $p$ and $q$, $\mathcal{N}(p)$ denotes the neighborhood of $p$, and $n$ denotes the number of nodes in the graph.

Next, the graph exploration algorithms for the dense graph $\mathcal{G}_d$ and the  sparse graph $\mathcal{G}_s$ are designed based on \textit{breadth first search} (BFS) and \textit{depth first search} (DFS), respectively \cite{networkintroduction}.  
This design is based on the properties that i) BFS can avoid visiting a node twice in one exploration, and ii) DFS can explore a graph as far as possible along a branch before backtracking. Therefore, given a collection of starting nodes, we accomplish the graph exploration by assessing each node's random walk probability, designed for DSF or BFS purpose. 

%
The random walk probability $\mathcal{P}_{p,q}$ of visiting a node $q$ from its adjacent node $p$ is defined as 
\begin{equation}
    \mathcal{P}_{p,q} = \frac{w_{p,q}}{\sum_{q\in \mathcal{N}(p)} w_{p,q}} \mbox{,}
    \label{eq:RWprob}
\end{equation}
where $w_{p,q}$ is the visiting weight from $p$ to $q$, the visiting weight $w^d_{p,q}$ for the dense graph is obtained in (\ref{eq:wpq_dense}), and the weight $w^s_{p,q}$ for the sparse graph is defined in (\ref{eq:wpq_sparse}). 
\begin{equation}
w^d_{p,q} = S_{\cos}(p,q) + \frac{1}{1+\log(d_q)}  \sum_{r\in\mathcal{N}_{p,q}} S(p,r) S(q,r) \mbox{,}
    \label{eq:wpq_dense}
\end{equation}
and
\begin{equation}
w^s_{p,q} = \frac{\lvert
    \Psi( \| \mathbf{f}_{\mathrm{AE}}(p) - \mathbf{f}_{\mathrm{AE}}(q) \|_2 )
    - 
    \Psi( \| \mathbf{f}_{\mathrm{SA}}(p) - \mathbf{f}_{\mathrm{SA}}(q) \|_2 )
    \rvert}{1+\log(d_q)} 
     \mbox{,}
    \label{eq:wpq_sparse}
\end{equation}
where $\mathcal{N}_{p,q}$ denotes the intersection of the one-ring-neighborhoods\footnote{The one-ring-neighborhood of a node $p$ is the set of all nodes connected with $p$ by an edge \cite{meyer2003discrete}.} of $p$ and $q$, $d_q$ is the degree of $q$, $\Psi(\cdot)$ is a min-max scaling function which maps an input value into [0, 1].
Moreover, $S(p,q)$ is the similarity score defined as the difference between i) the cosine similarity between nodes $p$ and $q$ and ii) the expected cosine similarity between any two nodes in $\mathcal{N}_{p,q}$, that is, 
\begin{equation}
    S(p,q) = S_{\cos}(p,q) - \frac{1}{C(|\mathcal{N}_{p,q}|,2)}\sum_{s,t\in \mathcal{N}_{p,q}}S_{\cos}(s,t) \mbox{,} 
\end{equation}
where $C(|\mathcal{N}_{p,q}|,2)$ denotes the number of 2-combinations of elements in $\mathcal{N}_{p,q}$. $S_{\cos}(p,q)$ is the cosine similarity between the latent features extracted by SA-LithoNet as follows: 
\begin{equation}
    S_{\cos}(p, q) = ( \mathbf f_{\mathrm{SA}}(p) \cdot \mathbf f_{\mathrm{SA}}(q)) / (|\mathbf f_{\mathrm{SA}}(p)| \cdot |\mathbf f_{\mathrm{SA}}(q)|)  \mbox{.} 
\end{equation}

Concisely, $w^s_{p,q}$ encourages visiting an adjacent $q$ with a distinct feature from $p$ for performing DFS on a sparse graph, whereas $w^d_{p,q}$ gives a larger weight to $q$ with a similar feature to $p$'s for performing BFS.

\begin{algorithm}[t!]
  \caption{  One-Time Sampling $\mathcal{S}_\mathrm{out}=\mathrm{OTS}(\mathcal{G}_\mathrm{in}, N_\mathrm{s}, \mathcal{M})$
  }
  \label{algo:vistinggraph}
  \begin{algorithmic}[1]
    \Require  
      Graph $\mathcal{G}_\mathrm{in}(\mathcal{V}, \mathcal{A})$, where $\mathcal{A}$ and $\mathcal{V}$ are respectively the adjacency matrix and node set of $\mathcal{G}_\mathrm{in}$;
      Required number of seed samples $N_\mathrm{s}$;
      Number of epochs $\mathcal{M}$. 
    \Ensure 
      Sampling set $\mathcal{S}_\mathrm{out}$;
    \State Evaluate an initial \textit{seed sore} for each node in $\mathcal{G}_\mathrm{in}$ based on (\ref{eq:eigencentral}) or (\ref{eq:closeness}); 
    \State Take nodes with the Top-$N_\mathrm{s}$ largest initial scores to form the starting-seed set $\mathcal{S}_\mathrm{seed}$;
    \For {all the $i$-th node $n_i \in \mathcal{S}_\mathrm{seed}$}
      \For {$epoch<\mathcal{M}$}
        \State $k \gets i$; 
        \For {$step < N_\mathrm{s}$}
          \State Update all $w_{k,j}$ via  (\ref{eq:wpq_dense}) or 
          (\ref{eq:wpq_sparse});
          \State Visit an adjacent $n_j$ randomly based on (\ref{eq:RWprob});
          \State $k \gets j$;
        \EndFor
      \EndFor
      \State For each node in $\mathcal{G}_\mathrm{in}$, total the number of visits;
      \State Select nodes of Top-$N_\mathrm{s}$ visits to form $\mathcal{S}_\mathrm{out}$; 
    \EndFor \\
    \Return $\mathcal{S}_\mathrm{out}$;
  \end{algorithmic}
\end{algorithm}

\subsection{Incremental Sampling}

Unlike the one-time sampling strategy, we further devise an incremental sampling method to split the total resource budget of fabricating unseen layout patterns and taking the corresponding SEM images into a few smaller fine-tuning datasets. In this way, we iteratively update a pretrained pre-simulation model to extend its generalization ability. To this end, the fine-tuning dataset selected in the $i$-th iteration should be able to  best update the pre-simulation model fine-tuned on the ($i-1$)-th fine-tuning dataset. 
 
The proposed incremental sampling method, taking the aforementioned one-time sampling method as its backbone, is an iterative routine with a stop-criterion function measuring the difference between the knowledge learned from two successive iterations under a resource budget.
As depicted in Algorithm \ref{alg:incresampling}, the main idea of our incremental sampling method is to re-rank the sampling priorities of unselected samples in the unseen-pattern pool after each sampling iteration with the aid of a meticulously-designed node attribute \textit{Informativeness-score}.

\noindent $\bullet$ \textbf{Informativeness-Score}: 
Assuming each sample carries a certain amount of information, say, information volume \cite{cui2019class}, the Informativeness-Score ($\mathcal{I}$-Score) $\mathcal{I}$ aims to assess the information volume carried by a selected sample in the feature space. 
Since the information volume covered by a frequently-visited node may usually be shared by its neighboring nodes, to avoid acquiring redundant information, the sampling priority of a frequently-visited node should be lower, and vice versa.
Therefore, we first take selected unseen samples as the starting nodes $n_i \in \mathcal{S}^{(l)}$, then evaluate the tendency of individual unselected nodes being visited by random walk, and finally evaluate the $\mathcal{I}$-Score $\mathcal{I}$ based on the tendency values. Algorithm \ref{alg:Learnedscore} shows the pseudo-code for evaluating the $\mathcal{I}$-Score. Note that i) $\mathcal{I}$ is a vector whose $k$-th entry $\mathcal{I}(k)$ denotes the $\mathcal{I}$-Score of the $k$-th node, and ii) in Algorithm \ref{alg:Learnedscore}, $\mathcal{I}_C^{(l)} = \mathcal{I}_C^{(l-1)} + \mathcal{I}^{(l)}$ denotes the vector whose entries record the cumulative $\mathcal{I}$-Score of individual nodes in the $l$-th iteration. 

\noindent $\bullet$ \textbf{Budget}: 
We exploit a variable $\mathcal{B}$, denoting \textit{budget}, to bound the maximal total visiting distance. This parameter is used to model the maximal information volume a starting node possesses, so $\mathcal{B}$ is set to be $\bm{k}$ used to construct our $\bm{k}$-NN graph. This design enables the Algorithm \ref{alg:Learnedscore} to visit at least $k$ nodes while evaluating the $\mathcal{I}$-Score.

\noindent $\bullet$ \textbf{Step--Cost}: 
We evaluate the \textit{cost} per move from $p$ to its neighbor $q$ based on the distance between autoencoder features and the ratio of graph densities between two nodes as follows:
\begin{equation}
\mathrm{Cost}(p,q)= \|\mathbf f_\mathrm{AE}(p) - \mathbf f_\mathrm{AE}(q) \|_2 \cdot\frac{\hat{\mathcal{D}}_\mathrm{SA}(q)}{\hat{\mathcal{D}}_\mathrm{SA}(p)} \mbox{,}
\label{eq:stepcost}
\end{equation}
where
\begin{equation}
  \hat{\mathcal{D}}_\mathrm{SA}(n_j)=
  \mathcal{D}_\mathrm{SA}(q)-\mathcal{A}(p,q) \cdot \mathcal{D}_\mathrm{SA}(p) \mbox{,}
\label{eq:D_litho}
\end{equation}
and
\begin{equation}
  \mathcal{D}_\mathrm{SA}(q) = \frac{\sum_p \mathcal{A}(p,q) \cdot \exp\big(-
  \frac{\|\mathbf f_\mathrm{SA}(p) - \mathbf f_\mathrm{SA}(q)\|_2}{2\sigma^2} \big)}{\sum_p \mathcal{A}(p,q)} \mbox{,}
\label{eq:Dhat_litho}
\end{equation}
where $\mathcal{D}_\mathrm{SA}$ denotes the graph density \cite{ebert2012ralf} measuring how close on average a node approaches its neighbors in the feature space spanned by $\mathbf f_\mathrm{SA}$, and $\hat{\mathcal{D}}_\mathrm{SA}$ prevents the same dense region from being selected redundantly by subtracting the weight of the destination node $q$ by the weight of the starting node $p$ of the current step. 


\noindent $\bullet$ \textbf{Tendency Weight}:
The tendency weight used to derive the random-walk probability of the $l$-th sampling iteration is given by 
\begin{equation}
w^{d,(l)}_{p,q} = S_{\cos}(p,q) +  \hat{\mathcal{I}}^{(l)}_{p,q}  \sum_{t\in\mathcal{N}_{p,q}} S(p,t)  S(q,t) \mbox{,}
    \label{eq:wpq_dense_incre}
\end{equation}
and
\begin{equation}
w^{s,(l)}_{p,q} =  \hat{\mathcal{I}}^{(l)}_{p,q} \cdot \lvert
    \Psi( \|\mathbf f_\mathrm{AE}(p) - \mathbf f_\mathrm{AE}(q) \|_2 )
    - 
    \Psi( \|\mathbf f_\mathrm{SA}(p) - \mathbf f_\mathrm{SA}(q) \|_2 )
    \rvert 
     \mbox{,}
     \label{eq:wpq_sparse_incre}
\end{equation}
where 
\begin{equation}
    \hat{\mathcal{I}}^{(l)}_{p,q} = \frac{\mathcal{I}_C^{(l)}(q)}{\sum_{t\in \mathcal{N}_{p,q}} \mathcal{I}_C^{(l)}(t)}
\end{equation}

Note that these two equations are similar to (\ref{eq:wpq_dense}) and (\ref{eq:wpq_sparse}) but use a different factor $\hat{\mathcal{I}}^{(l)}_{p,q}$ to balance the influence on the $t$-th node brought by the local neighborhood.

\noindent $\bullet$ \textbf{Stop Criterion}:
The stop criterion aims to check if 
the selected nodes represent the data graph well. 
This criterion implies that i) all samples on the graph can be equally visited by random walk, and ii) an additional batch of sampling cannot increase the normalized cumulative $\mathcal{I}$-Score of each node.
Hence, the stop criterion 
is defined as follows:
\begin{equation}
   \mathcal{L}_\mathrm{inc}^{(l)} = -\sum^{}_{k=1}{\|\bar{\mathcal{I}}_C^{(l)}(k)-\bar{\mathcal{I}}_C^{(l-1)}(k)\|\cdot\log_2(\bar{\mathcal{I}}_C^{(l-1)}(k)+\rho^{(l)})} \mbox{,}
\label{eq:Lincre}
\end{equation} 
where $\rho^{(l)}$ is the ratio of the number of selected samples after the $l$-th iteration to the number of total samples, and $\bar{\mathcal{I}}_C^{(l)}(i) = {\mathcal{I}_C^{(l)}(i)}/{\max_k\big( \mathcal{I}_C^{(l)}(k)\big)}$ is the normalized cumulative $\mathcal{I}$-Score.

\begin{algorithm}[t]
  \caption{Incremental Sampling}
  \label{alg:incresampling}
  \begin{algorithmic}[1]
    \Require  
      Input graph $\mathcal{G}_\mathrm{in}(\mathcal{V}_\mathrm{in}, \mathcal{A}_\mathrm{in})$;
      Number of epochs $\mathcal{M}$;
      Number of to-be-selected samples per batch $N_\mathrm{s}$;
    \Ensure 
      Sampling Set $\mathcal{S}_n$;
    \State $l\gets 1$; $\mathcal{I}_C^{(0)} \gets 0$;
    \State $\mathcal{S}_{n}$ = OneTimeSampling($\mathcal{G}_\mathrm{in}$, $N_\mathrm{s}$, $\mathcal{M}$);
    \State Update $\mathcal{L}_\mathrm{inc}$ via (\ref{eq:Lincre});
    \While {$\mathcal{L}_\mathrm{inc}>0$}
        \State Update $\mathcal{I}$-Score: $\mathcal{I}_C^{(l)}\gets \mathcal{I}_C^{(l-1)}+\mathcal{F}_\mathrm{IS}\big(\mathcal{G}_\mathrm{in}, \mathcal{S}^{(l)}, \mathcal{M} \big)$;
            \For {$epoch < \mathcal{M}$}
                \State $k \gets i$; 
                \For{$step < N_\mathrm{s}$}
                \State Update $w_{k,j}$ 
                via  (\ref{eq:wpq_dense_incre}) or (\ref{eq:wpq_sparse_incre});
                \State Visit an $n_j$ randomly based on 
                (\ref{eq:RWprob});
                \State $k \gets j$;
                \EndFor
            \State  For each node in $\mathcal{G}_\mathrm{in}$, total the number of visits;
            \EndFor
        \State $\mathcal{S}_\mathrm{tmp}=\{$Nodes with Top-$N_\mathrm{s}$ visits in $\mathcal{V}_\mathrm{in} \setminus \mathcal{S}^{(l)}\}$;
        \State $l \gets l+1$;
        \State $\mathcal{S}^{(l)} = \mathcal{S}^{(l-1)}\cup \mathcal{S}_\mathrm{tmp}$;
        \State Update $\mathcal{L}_\mathrm{inc}$;
    \EndWhile
    \State \Return $\mathcal{S}^{(l)}$;
  \end{algorithmic}
\end{algorithm}

\begin{algorithm}[t]
  \caption{Informativeness-Score $\mathbf{i}=\mathcal{F}_\mathrm{IS}\big(\mathcal{G}_\mathrm{in}, \mathcal{S}, \mathcal{M}\big)$}
  \label{alg:Learnedscore}
  \begin{algorithmic}[1]
    \Require  
      The input graph $\mathcal{G}_{in}$;
      Sampling set $\mathcal{S}$;
      Number of epochs $\mathcal{M}$;
    \Require 
      A global constant $\bm{k}$ used for constructing $\bm{k}$-NN graph;
    \Ensure 
      $\mathcal{I}$-score vector of nodes in $\mathcal{G}_\mathrm{in}$ after the $n$-th iteration.
     \For {All nodes $n_i$ in $\mathcal{S}$}
        \State  $s \gets i$;
        \State $\mathcal{B} \gets \bm{k}$;
        \While{ $\mathcal{B} \geq 0$ }
            \For {\textit{epoch} $< \mathcal{M}$}
                \State Update $w_{s,t}$ via (\ref{eq:wpq_dense_incre}) or (\ref{eq:wpq_sparse_incre});
                \State Visit an $n_t$ randomly based on (\ref{eq:RWprob});
                \State Evaluate $\mathrm{cost}(s,t)$ via (\ref{eq:stepcost});
                \State $\mathcal{B} \gets \mathcal{B} - \mathrm{cost}(s,t)$;
                \State $s \gets t$;
            \EndFor
        \EndWhile
    \EndFor
    \State Total the number of visits, \textit{i.e.}, $v_j$, to each $n_j$ in $\mathcal{G}_\mathrm{in}$; 
    \State \Return $\mathbf{i}= [v_1, v_2, \cdots, v_j, \cdots]$ 
  \end{algorithmic}
\end{algorithm}

	\section{Experimental Results}
	\label{sec:exp}
	\subsection{Dataset and Network Configuration}

\begin{figure}[t!]
    \begin{tabular}{ccc}
       \includegraphics[width=0.14\textwidth]{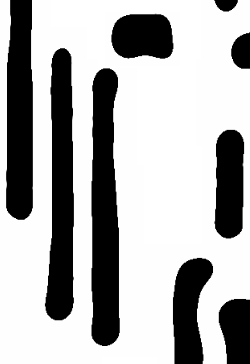}  
       & 
       \includegraphics[width=0.14\textwidth]{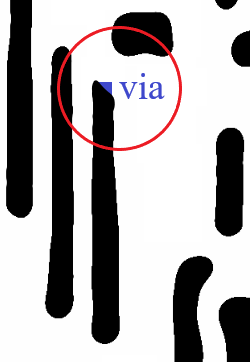}
       & 
       \includegraphics[width=0.14\textwidth]{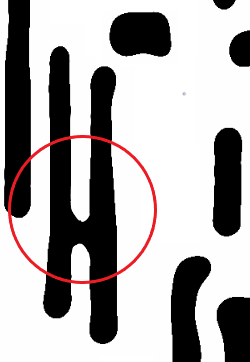}
       \\
       (a) Expectation & (b) Enclosure & (c) Bridge
    \end{tabular}
    \caption{Illustration of ``enclosure'' and ``bridge'' defect patterns in out datasets. (a) An expected defect-free SEM reference contour. (b) Enclosure pattern: enclosure means that the \textit{metal line} fails to enclose the \textit{via} due to contour shrinking after fabrication. (c) Bridge pattern: bridge means that unexpected connection between two metal lines occurs.
    }
    \label{fig:TwoDefect}
\end{figure}

Two datasets 
are used in our experiments. Both datasets comprise pair-wise image samples, each consisting of a  $1024\times1024$ layout pattern and a corresponding $1024\times1024$ binarized SEM image.
\textbf{Dataset-1} is used as \textit{seen} data, \textit{i.e.}, the training set, consisting of $1,036$ image pairs among which $886$ SEM images have ``enclosure'' patterns and the other $150$ have ``bridge'' patterns. 
Meanwhile, \textbf{Dataset-2}, the blind testing set, contains $1,000$ image pairs among which $150$ involve ``enclosure'' patterns and the remaining $850$ involve ``bridge'' patterns. 
Some examples of enclosure and bridge patterns are illustrated in Fig.~\ref{fig:TwoDefect}.
With this setting, we assumes  that the enclosure patterns in \textbf{Dataset-1} to be regular ones and the bridge patterns tend to be novelties.
Consequently, a successful novelty detection scheme should rate the bridge patterns in \textbf{Dataset-2}  with higher glocal novelty scores.
%

\begin{table}[]
    \caption{Architecture of the autoencoder used in our method}
    \centering
    \begin{tabular}{|c|c|c|}
    \hline
        \multicolumn{3}{|c|}{\footnotesize Encoder}\\
        \cline{1-3}
       \footnotesize  Layer & \footnotesize Filter & \footnotesize Output Size\\
        \cline{2-3}
        & \footnotesize $k\times k, s$ &\footnotesize $H\times W\times C$\\
        \hline
        \footnotesize Input & -- &\footnotesize  $1024\times1024\times1$\\
        \hline
        \footnotesize Conv-BN-ReLU & \footnotesize $5\times5,2$ & \footnotesize $512\times512\times32$\\
        \hline
        \footnotesize Conv-BN-ReLU & \footnotesize $5\times5,2$ & \footnotesize $256\times256\times64$\\
        \hline
        \footnotesize Conv-BN-ReLU & \footnotesize $5\times5,2$ & \footnotesize $128\times128\times128$\\
        \hline
        \footnotesize Conv-BN-ReLU & \footnotesize $5\times5,2$ & \footnotesize $64\times64\times128$\\
        \hline
        \hline
        \multicolumn{3}{|c|}{\footnotesize Decoder}\\ 
        \cline{1-3}
        \footnotesize Upsample & \footnotesize $2\times2,1$ & \footnotesize $128\times128\times128$\\
        \hline
        \footnotesize Conv-BN-LReLU & \footnotesize $5\times5,1$ & \footnotesize $128\times128\times128$\\
        \hline
        \footnotesize Upsample & \footnotesize $2\times2,1$ & \footnotesize $256\times256\times64$\\
        \hline
        \footnotesize Conv-BN-LReLU & \footnotesize $5\times5,1$ & \footnotesize $256\times256\times64$\\
        \hline
        \footnotesize Upsample & \footnotesize $2\times2,1$ & \footnotesize $512\times512\times32$\\
        \hline
        \footnotesize Conv-BN-LReLU & \footnotesize $5\times5,1$ & \footnotesize $512\times512\times32$\\
        \hline
        \footnotesize Upsample & \footnotesize $2\times2,1$ & \footnotesize $1024\times1024\times32$\\
        \hline
        \footnotesize Conv-BN-Sigmoid & \footnotesize $5\times5,1$ & \footnotesize $1024\times1024\times1$\\
        \hline
    \end{tabular}
    \label{tab:ae_architecture}
\end{table}

%

Both the auto-encoder and SA-LithoNet described in Fig.~\ref{fig_2} are pretrained on \textbf{Dataset-1}. For SA-LithoNet, we adopt the same LithoNet architecture and train it with the same settings used in~\cite{shao2021ic}. Table~\ref{tab:ae_architecture} shows the architecture of our auto-encoder, that is trained for $10$ epochs via the mean-squared-error (MSE) loss with a learning rate of $0.005$ and a batch-size of $1$.


We conduct two experiment sets to verify the effectiveness of our method. The first set validates whether our novelty detection scheme can accurately identify novel layout patterns, and the second evaluates the effectiveness of our sampling methods in selecting representative novel patterns for updating a pretrained pre-simulation model like LithoNet.

\subsection{Layout Novelty Detection}
\label{subsec:402}

In order to show the effectiveness of our layout novelty detection algorithm, we first verify the stability and robustness of our supervised MIGNA method, and then use the MIGNA results as the golden references to evaluate the accuracy of our global-local (glocal) layout novelty scoring.  

We use the AUC (Area Under the Curve) score  of the ROC (Receiver Operating Characteristic) curve as the objective evaluation metric. The higher the AUC score is, the more accurate the predictions are. Table~\ref{tab:different_tau} compares the AUC scores of the detection results with different novelty detection methods based on the MIGNA-annotated references (see  Sec.~\ref{subsec:301}) listed in the left three columns. Here, $\tau$ denotes the threshold of anomaly patches for assessing the layout novelty, and the number of layouts classified as novelty in \textbf{Dataset-2} decreases with $\tau$. 
The proposed \textbf{SA-Glocal novelty scoring}  outperforms the SA-LithoNet-based local scoring and autoencoder-based global scoring for all $\tau$ settings. 

\begin{table}[t]
    \caption{Comparison of AUC scores with four different methods under different settings of $\tau$ for assessing a novelty layout, where the best results are indicated in bold}
    \centering
    \footnotesize
    \begin{tabular}{|c||c|c||c|c|c|}
    \hline
         & \multicolumn{2}{c||}{MIGNA annotations} & \multicolumn{3}{c|}{AUC scores} \\
        \cline{2-6}
         $\tau$ & \# \textcolor{blue}{normal} & \# \textcolor{red}{novel}  & SA-Litho  & Autoencoder  & \textbf{Ours}\\
         &  &   & (Local) & (Global) & \textbf{(Glocal)}\\
         \hline
        3 & 299 & 701 &  0.825 & 0.684 & \textbf{0.862}\\
        \hline
        4 & 383 & 617 &  0.805 & 0.737 & \textbf{0.861}\\
        \hline
        5 & 449 & 551 &  0.744 & 0.749 & \textbf{0.846}\\
        \hline
        6 & 559 & 441 &  0.683 & 0.683 & \textbf{0.756}\\
        \hline
        7 & 655 & 345 &  0.624 & 0.609 & \textbf{0.676}\\
        \hline
    \end{tabular}
    \label{tab:different_tau}
\end{table}

\begin{figure}[!t]
    \centering
    \includegraphics[width=0.40\textwidth]{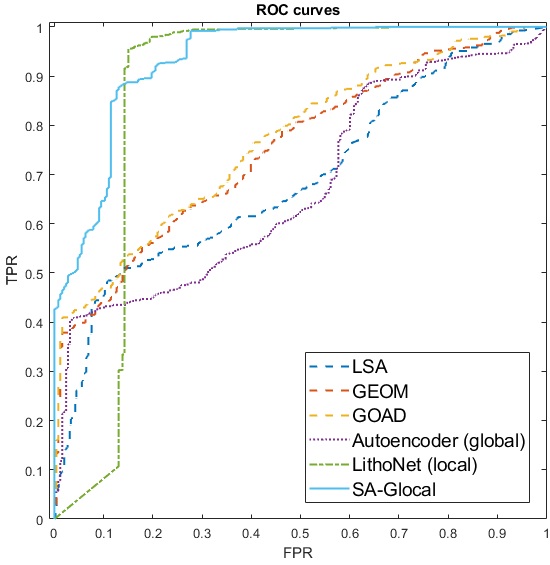}\\
 \vspace{-0.2cm}
    \caption{ROC curves on \textbf{Dataset-2} with different novelty detection schemes, including the proposed SA-Glocal score, the SA-LithoNet-based local novelty score, and the autoencoder-based global novelty score, and three representative ones: LSA~\cite{abati2019latent}, GEOM~\cite{golan2018deep}, and GOAD~\cite{bergman2020classification}.} 
    \label{fig:RocDistri}
\end{figure}

Fig.~\ref{fig:RocDistri} shows the ROC performances on \textbf{Dataset-2} with different novelty detection methods, including our methods (autoencoder-based, LithoNet-based, and SA-Glocal) and three state-of-the-art novelty detection approaches:  LSA~\cite{abati2019latent}, GEOM~\cite{golan2018deep}, and GOAD~\cite{bergman2020classification}. In Fig.~\ref{fig:RocDistri}, the MIGNA-annotated labels  are used as pseudo ground-truths to calculate the true positive rates (TPRs) and false positive rates (FPRs).
The ROC curves demonstrate that the LithoNet-based local novelty scoring and the autoencoder-based global novelty scoring are complementary to each other, where the former detects much more novelties at low FPRs while the latter can detect almost all novelties at about $20\%$ FPR. Consequently, they can be combined together to boost the performance of novelty detection, as shown via the SA-Glocal curve. 
 Table~\ref{tab:comp_novel} lists the AUC scores with the six schemes, showing that the proposed \textbf{SA-Glocal} novelty scoring well beats all the others, achieving a significantly higher AUC score of $0.932$. 

\begin{figure*}
    \centering
    \includegraphics[width=0.85\textwidth]{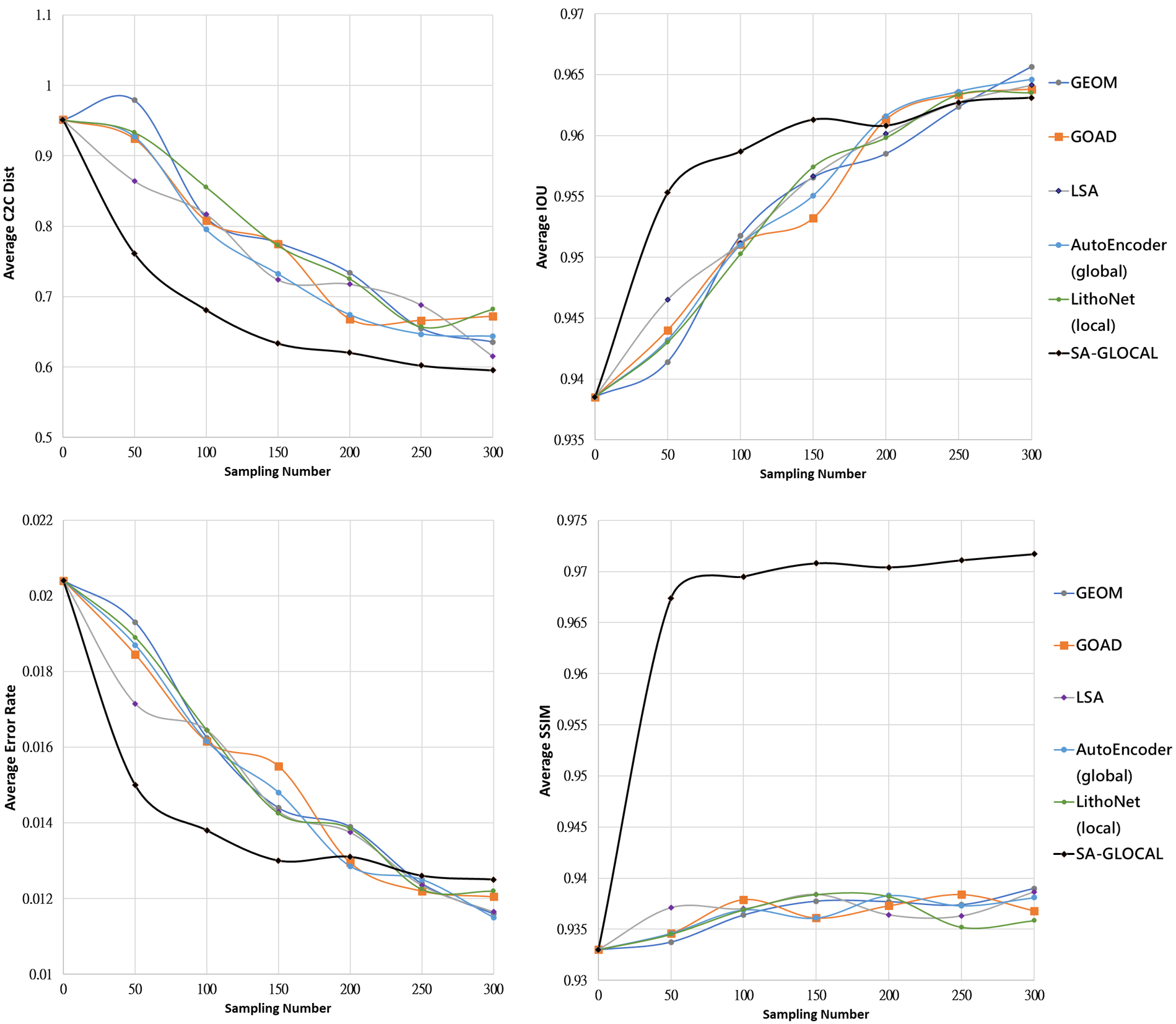}
    \caption{Inference performance of LithoNet fine-tuned on different finetune sets of different sizes (50, 100, 150, 200, 250, and 300 samples) randomly selected from the novel samples detected by a specific novelty detection scheme.}
    \label{figX:figX}
\end{figure*}

To validate the impacts of various novelty detection schemes on the performance of model update, we randomly select 50, 100, 150, 200, and 250 samples out of the novel patterns detected by each method, together with the original training set, to form the finetune sets of different sizes, and then use them to update the LithoNet model. Fig. \ref{figX:figX} compares the inference performances of different fine-tuned LithoNet models, where each point on a curve corresponds to a LithoNet updated by a finetune set containing randomly-selected novel samples detected by one specific novelty detection scheme. The horizontal axis indicates the amount of novel samples randomly picked into the finetune dataset. Note that we here adopt the same similarity metrics used in \cite{shao2021ic}, including C2C-distance (contour-to-contour distance) \cite{C2Cdist}, IOU (intersection over union), SSIM (structural similarity index measure) \cite{SSIM}, and pixel-error-rate, to evaluate  the performances of the LithoNet models updated on the various fine-tune sets. The results demonstrate that, for a pretrained LithoNet model, the novel samples detected by our SA-Glocal are significantly more informative than those detected by LSA, GOEM, and GOAD, making SA-Glocal outperform the competing methods in terms of all quality metrics for model update. Moreover,  Fig. \ref{figX:figX} also hints that SA-Glocal scoring is capable of being an active learning oracle because even a very limited amount of randomly-selected novel patterns identified by SA-Glocal can best fine-tune a pretrained LithoNet. 

Table~\ref{tab:comp_novel_ours} shows the ablation study of our novelty detection method. Here, LithoNet (OC-SVM) local scoring shows the baseline performance, \textit{i.e.}, an AUC value of 0.720, obtained by feeding the LithoNet latent codes of test layouts into the conventional  one-class SVM outlier detector~\cite{liu2013svdd}. 
The MC-SVM-based LithoNet scoring presented in Sec.~\ref{subsec:local} improves the AUC score to 0.744. 
Moreover, by combining the autoencoder global feature with the LithoNet local feature (\textit{i.e.}, the Glocal method), the AUC score significantly increases to 0.846. This demonstrates the effectiveness and the robustness of our Glocal design. Finally, the last three rows in Table~\ref{tab:comp_novel_ours} evidence that SA-LithoNet can further boost the representability of the latent feature, particularly making  the proposed SA-Glocal method achieve the best performance: 0.932 AUC score.


\begin{table}[t]
\small
    \caption{AUC scores of different layout novelty detection methods on \textbf{Dataset-2}, where the best and second-best results are respectively highlighted in bold and underline}
    \centering
    \begin{tabular}{|l|c|}
    \hline
        \footnotesize Method &\footnotesize AUC Score \\
        \hline 
        \footnotesize LSA~\cite{abati2019latent} & \footnotesize 0.690 \\
        \hline
        \footnotesize GEOM~\cite{golan2018deep} & \footnotesize 0.752 \\
        \hline
        \footnotesize GOAD~\cite{bergman2020classification} & \footnotesize \underline{0.768} \\ 
        \hline
        \footnotesize Autoencoder (Global) & \footnotesize 0.675 \\
        \hline
       \footnotesize LithoNet (Local) & \footnotesize {0.744} \\
        \hline
        \footnotesize \textbf{SA-Glocal (AE + SA-LithoNet)} & \footnotesize \textbf{0.932} \\   
    \hline     
    \end{tabular}
    \label{tab:comp_novel}
     \vspace{-0.2cm}
\end{table}

\begin{table}[t!]
\small
    \caption{Ablation study: the AUC scores of the proposed method and its variants on \textbf{Dataset-2}}
    \centering
    \begin{tabular}{|l|c|}
    \hline
       \footnotesize  Method & \footnotesize AUC Score \\
        \hline
        \footnotesize Autoencoder  & \footnotesize {0.675} \\
        \hline
        \footnotesize LithoNet (OC-SVM) & \footnotesize {0.720} \\
        \hline
        \footnotesize LithoNet (MC-SVM) & \footnotesize {0.744} \\
        \hline
        \footnotesize Glocal (AE + LithoNet) & \footnotesize {0.846} \\
        \hline
        \footnotesize SA-LithoNet (OC-SVM) & \footnotesize {0.857} \\
        \hline
        \footnotesize SA-LithoNet (MC-SVM) & \footnotesize {0.864} \\
        \hline
        \footnotesize \textbf{SA-Glocal (AE + SA-LithoNet)} & \footnotesize \textbf{0.932} \\   
    \hline     
    \end{tabular}
    \label{tab:comp_novel_ours}
\end{table}

\subsection{Performance Evaluation on Active-Learning Schemes}

The experiments reported herein are conducted by the following steps. 
First, for \textbf{Dataset-2}, we label those samples whose  \textbf{SA-Glocal} score $\Theta_\mathrm{novel} \geq -0.9$ as novelties, and regular patterns otherwise.
Second, we partition \textbf{Dataset-2} into two subsets: 
i) a finetune set $\mathcal{D}_\mathrm{FT}$ consisting of a subset $\mathcal{D}_\mathrm{FT}^r$ with randomly picked $300$ regular image pairs and a subset $\mathcal{D}_\mathrm{FT}^n$ with randomly picked $460$ novel pairs, and ii) a blind testing set $\mathcal{D}_\mathrm{test}$ comprising $120$ regular pairs and $120$ novel pairs. 
Third, we update the pretrained LithoNet individually on the finetune sets $\mathcal{D}_\mathrm{FT}$ together with the original training set, selected by different active learning strategies, and then evaluate their model performance on $\mathcal{D}_\mathrm{test}$. 

\begin{figure*}[!t]
    \centering
    \includegraphics[width=0.95\textwidth]{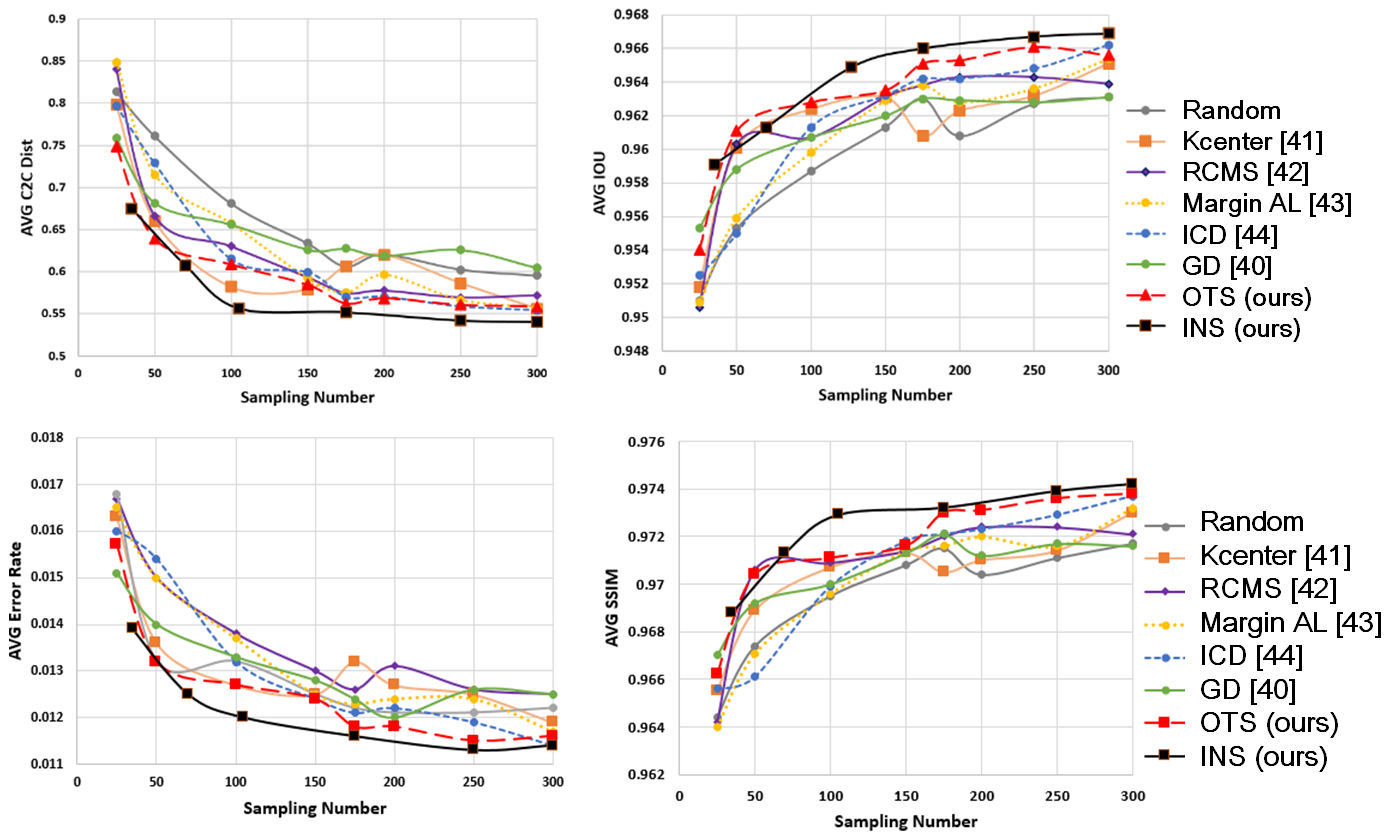}
    \caption{Inference performances of LithoNet fine-tuned on different finetune sets sampled by various active approaches from the detected novelties. Each plot shows the curves of one specific performance metric, including C2C-distance, SSIM, Error-rate, and IOU values.
    }
    \label{fig_all_c2c}
\end{figure*}
\begin{figure*}[!t]
    \centering
    \includegraphics[width=0.99\textwidth]{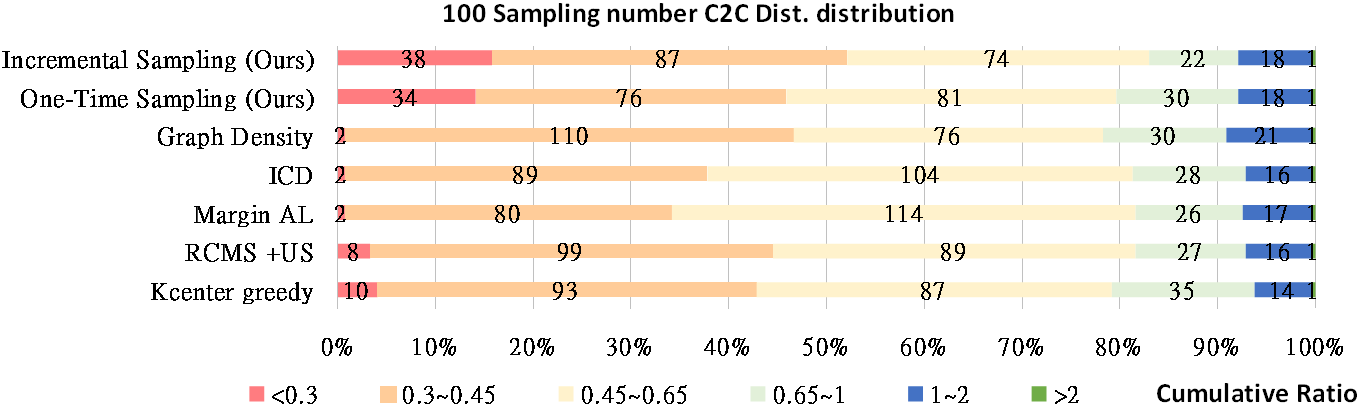}
    \caption{Breakdowns of inference performances of LithoNet models updated on different finetune sets, all containing $100$ samples. Each bar shows the breakdown of a specific range of C2C-distances obtained on the testing samples. 
    It is obvious that the LithoNet models fine-tuned on the $\mathcal{D}^n_\mathrm{FT}$, formed by our incremental sampling scheme or our one-time sampling schemes, have best generalization ability since most testing samples result in C2C-distances smaller than $0.65$ pixels through these two LithoNet models.
    }
    \label{fig_15_c2c}
\end{figure*}

Fig.~\ref{fig_all_c2c} compares the performance with various active learning strategies, where each point on a curve corresponds to a different subset of $\mathcal{D}_\mathrm{FT}$. We compare the proposed one-time sampling (\textbf{OTS}) and incremental sampling (\textbf{INS}) methods with random sampling and existing \textbf{active/graph} sampling methods, including K-center greedy (Kcenter)~\cite{sener2018active}, RCMS with uncertainly sampling (RCMS)~\cite{xu2003representative}, Margin AL \cite{zhou2014improved}, informative cluster diversity (ICD)~\cite{paul2016efficient}, and graph density~\cite{ebert2012ralf}. The horizontal axis in Fig.~\ref{fig_all_c2c} indicates the 
amount of novel samples selected into the finetune set.
 Moreover, Fig.~\ref{fig_15_c2c} shows the breakdowns of C2C distance ranges obtained on the testing dataset $\mathcal{D}_\mathrm{test}$ with different LithoNet models fine-tuned on different $100$-sample\footnote{The sampling amount of the proposed incremental sampling method (INS) cannot be assigned in advance and is determined during run-time. The number of samples closest to 100 calculated by INS is 105. } fine-tune sets selected by different sampling methods, respectively. 
The comparison shows that the two LithoNet models, respectively fine-tuned on the two $100$-sample fine-tune sets selected by our proposed one-time sampling  and incremental sampling schemes, are best improved. Specifically, the C2C-distances of more than $45\%$ of testing samples are less than 0.45 pixel. In contrast, the two LithoNets, fine-tuned on sample sets selected by 
RCMS~\cite{xu2003representative} and Graph Density (GD)~\cite{ebert2012ralf}, achieve the closest performances to the OTS and INS fine-tuned models. However, they result in much fewer test samples within the least  C2C-distances range (\textit{i.e.}, $[0, 0.3$]) than our methods', resulting in the performance differences illustrated in Fig.~\ref{fig_all_c2c}.

We can conclude from  Fig.~\ref{fig_all_c2c} and Fig.~\ref{fig_15_c2c} the following observations. First, the novelties detected by the Glocal novelty scoring are beneficial for updating a pretrained LithoNet since i) the C2C-distance and error-rate decrease, and ii) the SSIM and IOU scores increase with the number of selected samples. This observation is reasonable because the benefit of adding new data points to a training dataset will diminish if these new data points are significantly similar to 
existing samples in the training dataset, 
as revealed in~\cite{cui2019class}. Second, while the proposed one-time sampling (OTS) method outperforms the competing methods in all aspects, the proposed incremental sampling (INS) method has best cures and can reach the performance plateau when only sampling 105/460 of the data. This means that our  novelty detection together with graph sampling can effectively accomplish the goal of active learning from novel layout clips.

	\section{Conclusions}
	\label{sec:conclusion}
	
In this paper, we proposed a deep learning-based layout novelty detection method that can work in the absence of ground-truth SEM images. The proposed method architecturally consists of two subnetworks, a pretrained autoencoder and a pretrained layout-to-SEM simulator. The former subnetwork learns to capture global shape structures of training (layout) samples so that it can be used to derive the autoencoder-based global novelty score. Besides, the latter subnetwork aims to extract a latent code representing the fabrication-induced local shape deformation of a given layout so that the extracted latent code can be used to evaluate an attention-guided local novelty score. These two novelty scores together form the proposed \textit{Glocal layout novelty measure}. We have also proposed two graph sampling-based active-learning strategies, one-time sampling and incremental sampling, to  select a much reduced set of representative layouts most worthy of further fabrication for acquiring the ground-truth SEM images, in an on-a-budget environment.
Our experimental results demonstrate that the proposed method can detect novel layout patterns effectively, and the identified layout novelties can be used to improve the generalization capability of a learning-based layout-to-SEM pre-simulation model.

	
	
	\bibliographystyle{IEEEtran}
	\bibliography{sec10_reference}

\begin{thebibliography}{10}
\providecommand{\url}[1]{#1}
\csname url@samestyle\endcsname
\providecommand{\newblock}{\relax}
\providecommand{\bibinfo}[2]{#2}
\providecommand{\BIBentrySTDinterwordspacing}{\spaceskip=0pt\relax}
\providecommand{\BIBentryALTinterwordstretchfactor}{4}
\providecommand{\BIBentryALTinterwordspacing}{\spaceskip=\fontdimen2\font plus
\BIBentryALTinterwordstretchfactor\fontdimen3\font minus
  \fontdimen4\font\relax}
\providecommand{\BIBforeignlanguage}[2]{{%
\expandafter\ifx\csname l@#1\endcsname\relax
\typeout{** WARNING: IEEEtran.bst: No hyphenation pattern has been}%
\typeout{** loaded for the language `#1'. Using the pattern for}%
\typeout{** the default language instead.}%
\else
\language=\csname l@#1\endcsname
\fi
#2}}
\providecommand{\BIBdecl}{\relax}
\BIBdecl

\bibitem{yang2019gan}
H.~Yang, S.~Li, Z.~Deng, Y.~Ma, B.~Yu, and E.~F. Young, ``{GAN-OPC}: Mask
  optimization with lithography-guided generative adversarial nets,''
  \emph{IEEE Trans. Comput.-Aided Design Integr. Circuits Syst.}, vol.~39,
  no.~10, pp. 2822--2834, 2019.

\bibitem{ye2019lithogan}
W.~Ye, M.~B. Alawieh, Y.~Lin, and D.~Z. Pan, ``Lithogan: End-to-end lithography
  modeling with generative adversarial networks,'' in \emph{Proc. ACM/IEEE
  Design Autom. Conf.}, 2019, pp. 1--6.

\bibitem{shao2021ic}
H.-C. Shao, C.-Y. Peng, J.-R. Wu, C.-W. Lin, S.-Y. Fang, P.-Y. Tsai, and Y.-H.
  Liu, ``From ic layout to die photo: {A CNN}-based data-driven approach,''
  \emph{IEEE Trans. Comput.-Aided Design Integr. Circuits Syst.}, vol.~40,
  no.~5, pp. 957--970, May 2021.

\bibitem{pimentel2014review}
M.~Pimentel, D.~Clifton, L.~Clifton, and L.~Tarassenko, ``A review of novelty
  detection,'' \emph{Signal Process.}, vol.~99, pp. 215--249, 2014.

\bibitem{salehi2021unified}
M.~Salehi, H.~Mirzaei, D.~Hendrycks, Y.~Li, M.~Rohban, and M.~Sabokrou, ``A
  unified survey on anomaly, novelty, open-set, and out-of-distribution
  detection: Solutions and future challenges,'' \emph{arXiv preprint
  arXiv:2110.14051}, 2021.

\bibitem{liu2013svdd}
B.~Liu, Y.~Xiao, L.~Cao, Z.~Hao, and F.~Deng, ``Svdd-based outlier detection on
  uncertain data,'' \emph{Knowledge and Inf. Syst.}, vol.~34, no.~3, pp.
  597--618, 2013.

\bibitem{terrell1992variable}
G.~R. Terrell and D.~W. Scott, ``Variable kernel density estimation,''
  \emph{Annals of Statistics}, pp. 1236--1265, 1992.

\bibitem{japkowicz1995novelty}
N.~Japkowicz, C.~Myers, M.~Gluck \emph{et~al.}, ``A novelty detection approach
  to classification,'' in \emph{Proc. Int. Joint Conf. Artif. Intell.}, vol.~1,
  1995, pp. 518--523.

\bibitem{miljkovic2010review}
D.~Miljkovi{\'c}, ``Review of novelty detection methods,'' in \emph{Proc. Int.
  Convention MIPRO}, 2010, pp. 593--598.

\bibitem{calderara2011detecting}
S.~Calderara, U.~Heinemann, A.~Prati, R.~Cucchiara, and N.~Tishby, ``Detecting
  anomalies in people’s trajectories using spectral graph analysis,''
  \emph{Comput. Vis. Image Understand.}, vol. 115, no.~8, pp. 1099--1111, 2011.

\bibitem{mathieu2015masked}
M.~Mathieu, ``Masked autoencoder for distribution estimation,'' 2015.

\bibitem{schlegl2017unsupervised}
T.~Schlegl, P.~Seeb{\"o}ck, S.~M. Waldstein, U.~Schmidt-Erfurth, and G.~Langs,
  ``Unsupervised anomaly detection with generative adversarial networks to
  guide marker discovery,'' in \emph{Proc. Int. Conf. Inf. Process. Med.
  Imag.}, 2017, pp. 146--157.

\bibitem{fan2020anomalydae}
H.~Fan, F.~Zhang, and Z.~Li, ``Anomalydae: Dual autoencoder for anomaly
  detection on attributed networks,'' in \emph{Proc. IEEE Int. Conf. Acoustics
  Speech Signal Process.}, 2020, pp. 5685--5689.

\bibitem{wang2020one}
X.~Wang, B.~Jin, Y.~Du, P.~Cui, and Y.~Yang, ``One-class graph neural networks
  for anomaly detection in attributed networks,'' \emph{arXiv preprint
  arXiv:2002.09594}, 2020.

\bibitem{an2015variational}
J.~An and S.~Cho, ``Variational autoencoder based anomaly detection using
  reconstruction probability,'' \emph{Special Lecture on IE}, vol.~2, no.~1,
  pp. 1--18, 2015.

\bibitem{kliger2018novelty}
M.~Kliger and S.~Fleishman, ``Novelty detection with gan,'' \emph{arXiv
  preprint arXiv:1802.10560}, 2018.

\bibitem{abati2019latent}
D.~Abati, A.~Porrello, S.~Calderara, and R.~Cucchiara, ``Latent space
  autoregression for novelty detection,'' in \emph{Proc. IEEE/CVF Conf. Comput.
  Vis. Pattern Recognit.}, 2019, pp. 481--490.

\bibitem{sabokrou2018adversarially}
M.~Sabokrou, M.~Khalooei, M.~Fathy, and E.~Adeli, ``Adversarially learned
  one-class classifier for novelty detection,'' in \emph{Proc. IEEE/CVF Conf.
  Comput. Vis. Pattern Recognit.}, 2018, pp. 3379--3388.

\bibitem{perera2019ocgan}
P.~Perera, R.~Nallapati, and B.~Xiang, ``Ocgan: One-class novelty detection
  using gans with constrained latent representations,'' in \emph{Proc. IEEE/CVF
  Conf. Comput. Vis. Pattern Recognit.}, 2019, pp. 2898--2906.

\bibitem{sung2019difference}
Y.~L. Sung, S.-H. Hsieh, S.-C. Pei, and C.-S. Lu, ``Difference-seeking
  generative adversarial network--unseen sample generation,'' in \emph{Proc.
  Int. Conf. Learn. Rep.}, 2019.

\bibitem{pidhorskyi2018generative}
S.~Pidhorskyi, R.~Almohsen, D.~A. Adjeroh, and G.~Doretto, ``Generative
  probabilistic novelty detection with adversarial autoencoders,'' \emph{arXiv
  preprint arXiv:1807.02588}, 2018.

\bibitem{golan2018deep}
I.~Golan and R.~El-Yaniv, ``Deep anomaly detection using geometric
  transformations,'' \emph{arXiv preprint arXiv:1805.10917}, 2018.

\bibitem{bergman2020classification}
L.~Bergman and Y.~Hoshen, ``Classification-based anomaly detection for general
  data,'' \emph{arXiv preprint arXiv:2005.02359}, 2020.

\bibitem{tack2020csi}
J.~Tack, S.~Mo, J.~Jeong, and J.~Shin, ``Csi: Novelty detection via contrastive
  learning on distributionally shifted instances,'' \emph{arXiv preprint
  arXiv:2007.08176}, 2020.

\bibitem{settles2009active}
B.~Settles, ``Active learning literature survey,'' 2009.

\bibitem{lewis1995sequential}
D.~D. Lewis, ``A sequential algorithm for training text classifiers:
  Corrigendum and additional data,'' in \emph{ACM SIGIR Forum}, vol.~29, no.~2,
  1995, pp. 13--19.

\bibitem{tong2001support}
S.~Tong and D.~Koller, ``Support vector machine active learning with
  applications to text classification,'' \emph{J. Mach. Learn. Res.}, vol.~2,
  no. Nov, pp. 45--66, 2001.

\bibitem{lin2018data}
Y.~Lin, M.~Li, Y.~Watanabe, T.~Kimura, T.~Matsunawa, S.~Nojima, and D.~Z. Pan,
  ``Data efficient lithography modeling with transfer learning and active data
  selection,'' \emph{IEEE Trans. Comput.-Aided Design Integr. Circuits Syst.},
  vol.~38, no.~10, pp. 1900--1913, 2018.

\bibitem{zhuo2010active}
C.~Zhuo, K.~Agarwal, D.~Blaauw, and D.~Sylvester, ``Active learning framework
  for post-silicon variation extraction and test cost reduction,'' in
  \emph{Proc. IEEE/ACM Int. Conf. Comput.-Aided Design}.\hskip 1em plus 0.5em
  minus 0.4em\relax IEEE, 2010, pp. 508--515.

\bibitem{zhou2017fine}
Z.~Zhou, J.~Shin, L.~Zhang, S.~Gurudu, M.~Gotway, and J.~Liang, ``Fine-tuning
  convolutional neural networks for biomedical image analysis: actively and
  incrementally,'' in \emph{Proc. IEEE Conf. Comput. Vis. Pattern Recognit.},
  2017, pp. 7340--7351.

\bibitem{vaswani2017attention}
A.~Vaswani, N.~Shazeer, N.~Parmar, J.~Uszkoreit, L.~Jones, A.~N. Gomez,
  {\L}.~Kaiser, and I.~Polosukhin, ``Attention is all you need,'' in \emph{Adv.
  Neural Inf. Process. Syst.}, 2017, pp. 5998--6008.

\bibitem{dosovitskiy2020image}
A.~Dosovitskiy, L.~Beyer, A.~Kolesnikov, D.~Weissenborn, X.~Zhai,
  T.~Unterthiner, M.~Dehghani, M.~Minderer, G.~Heigold, S.~Gelly \emph{et~al.},
  ``An image is worth 16x16 words: Transformers for image recognition at
  scale,'' in \emph{Proc. Int. Conf. Learn. Rep.}, 2021.

\bibitem{wang2018nonlocal}
X.~Wang, R.~Girshick, A.~Gupta, and K.~He, ``Non-local neural networks,'' in
  \emph{Proc. IEEE/CVF Conf. Comput. Vis. Pattern Recognit.}, 2018, pp.
  7794--7803.

\bibitem{zhang2019self}
H.~Zhang, I.~Goodfellow, D.~Metaxas, and A.~Odena, ``Self-attention generative
  adversarial networks,'' in \emph{Proc. Int. Conf. Mach. Learn.}, 2019, pp.
  7354--7363.

\bibitem{li2003improving}
K.-L. Li, H.-K. Huang, S.-F. Tian, and W.~Xu, ``Improving one-class svm for
  anomaly detection,'' in \emph{Proc. Int. Conf. Mach. Learn. Cybern.}, vol.~5,
  pp. 3077--3081.

\bibitem{chang2013revisit}
W.-C. Chang, C.-P. Lee, and C.-J. Lin, ``A revisit to support vector data
  description,'' \emph{Dept. Comput. Sci., Nat. Taiwan Univ., Taipei, Taiwan,
  Tech. Rep}, 2013.

\bibitem{networkintroduction}
M.~Newman, \emph{Networks: An Introduction}.\hskip 1em plus 0.5em minus
  0.4em\relax Oxford University Press, 2010.

\bibitem{meyer2003discrete}
M.~Meyer, M.~Desbrun, P.~Schr{\"o}der, and A.~Barr, ``Discrete
  differential-geometry operators for triangulated 2-manifolds,'' in
  \emph{Visualization and mathematics III}, H.-C. Hege and K.~Polthier,
  Eds.\hskip 1em plus 0.5em minus 0.4em\relax Springer, 2003, pp. 35--57.

\bibitem{cui2019class}
Y.~Cui, M.~Jia, T.-Y. Lin, Y.~Song, and S.~Belongie, ``Class-balanced loss
  based on effective number of samples,'' in \emph{Proc. IEEE/CVF Conf. Comput.
  Vis. Pattern Recognit.}, 2019, pp. 9268--9277.

\bibitem{ebert2012ralf}
S.~Ebert, M.~Fritz, and B.~Schiele, ``Ralf: A reinforced active learning
  formulation for object class recognition,'' in \emph{Proc. IEEE/CVF Conf.
  Comput. Vis. Pattern Recognit.}, 2012, pp. 3626--3633.

\bibitem{C2Cdist}
H.-C. Shao, ``Contour-to-contour distance,''
  \url{https://www.mathworks.com/matlabcentral/fileexchange/75551-contour-to-contour-distance}.

\bibitem{SSIM}
Z.~Wang, A.~C. Bovik, H.~R. Sheikh, E.~P. Simoncelli \emph{et~al.}, ``Image
  quality assessment: from error visibility to structural similarity,''
  \emph{IEEE Trans. Image Process.}, vol.~13, no.~4, pp. 600--612, 2004.

\bibitem{sener2018active}
O.~Sener and S.~Savarese, ``Active learning for convolutional neural networks:
  A core-set approach,'' in \emph{Proc. Int. Conf. Learn. Rep.}, 2018.

\bibitem{xu2003representative}
Z.~Xu, K.~Yu, V.~Tresp, X.~Xu, and J.~Wang, ``Representative sampling for text
  classification using support vector machines,'' in \emph{Proc. European Conf.
  Inf. Retr.}, 2003, pp. 393--407.

\bibitem{zhou2014improved}
J.~Zhou and S.~Sun, ``Improved margin sampling for active learning,'' in
  \emph{Proc. Chinese Conf. Pattern Recognit.}, 2014, pp. 120--129.

\bibitem{paul2016efficient}
S.~Paul, J.~Bappy, and A.~Roy-Chowdhury, ``Efficient selection of informative
  and diverse training samples with applications in scene classification,'' in
  \emph{Proc. IEEE Inf. Conf. Image Process.}\hskip 1em plus 0.5em minus
  0.4em\relax IEEE, 2016, pp. 494--498.

\end{thebibliography}

	\vspace{-0.4in}
\begin{IEEEbiography}
[{\includegraphics[width=1in,height=1.25in,clip,keepaspectratio]{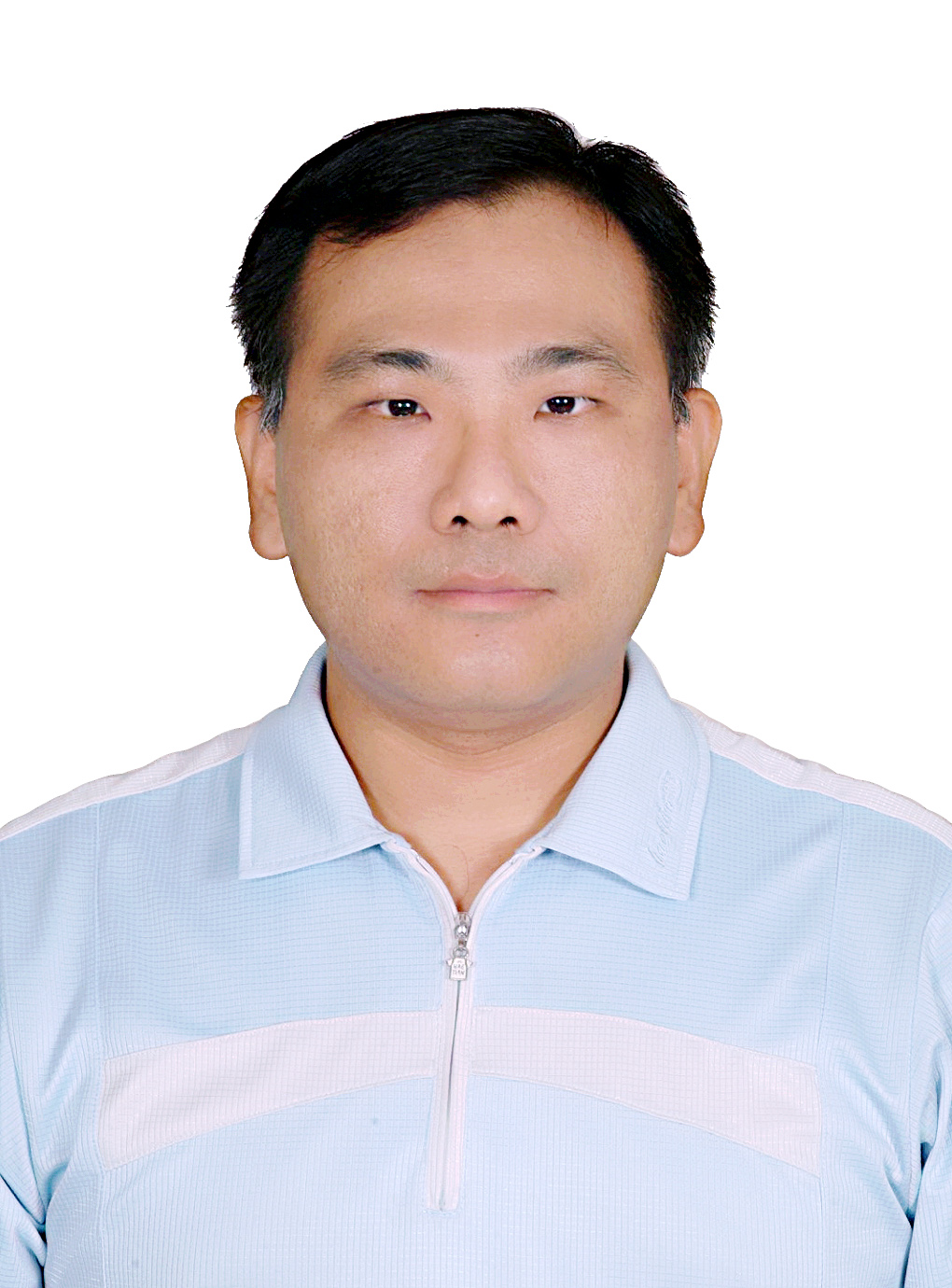}}]
{Hao-Chiang~Shao}
(Member, IEEE) received his Ph.D. degree in electrical engineering from National Tsing Hua University, Taiwan, in 2012. He has been an Assistant Professor with the Dept. Statistics and Information Science, Fu Jen Catholic University, Taiwan, since 2018. During 2012 to 2017, he was a postdoctoral researcher with the Institute of Information Science, Academia Sinica, involved in a series of \textit{Drosophila} brain research projects; in 2017--2018, he was an R\&D engineer with the Computational Intelligence Technology Center, Industrial Technology Research Institute, Taiwan, taking charges of DNN-based automated optical inspection (AOI) projects. His research interests include 2D+Z image atlasing, 3D mesh processing, big industrial image data analysis, and machine learning.
\end{IEEEbiography}

\vspace{-0.4in}
\begin{IEEEbiography}
	[{\includegraphics[width=1in,height=1.25in,clip,keepaspectratio] {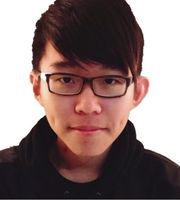}}]
	{Hsing-Lei Ping}
	 received his B.S. degree in Electronic engineering from National Chiao Tung University in 2019, and M.S. degrees in Electrical Engineering from National Tsing Hua University, in 2021.  
	He joined Phison Electronics Corp. as a software engineer since 2021.   His research interests lie in computer vision, machine learning, and visual analytics for IC design for manufacturability.
\end{IEEEbiography}

\vspace{-0.4in}
\begin{IEEEbiography}
	[{\includegraphics[width=1in,height=1.25in,clip,keepaspectratio] {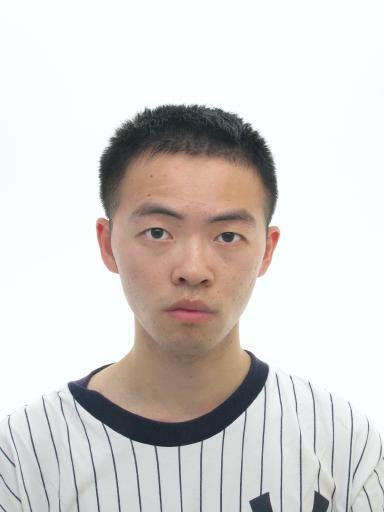}}]
	{Kuo-shiuan Chen}
	 received his B.S.degree in Power Mechanical Engineering and M.S. degree in Electrical engineering both from National Tsing Hua University  in 2019 and 2021, respectively.  
	He is currently working for Cadence Design Systems, Inc. (Cadence) as an engineer.   His research interests lie in computer vision, machine learning, and visual analytics for IC design for manufacturability.
\end{IEEEbiography}

\vspace{-0.4in}
\begin{IEEEbiography}[{\includegraphics[width=1in,height=1.25in,clip,keepaspectratio]{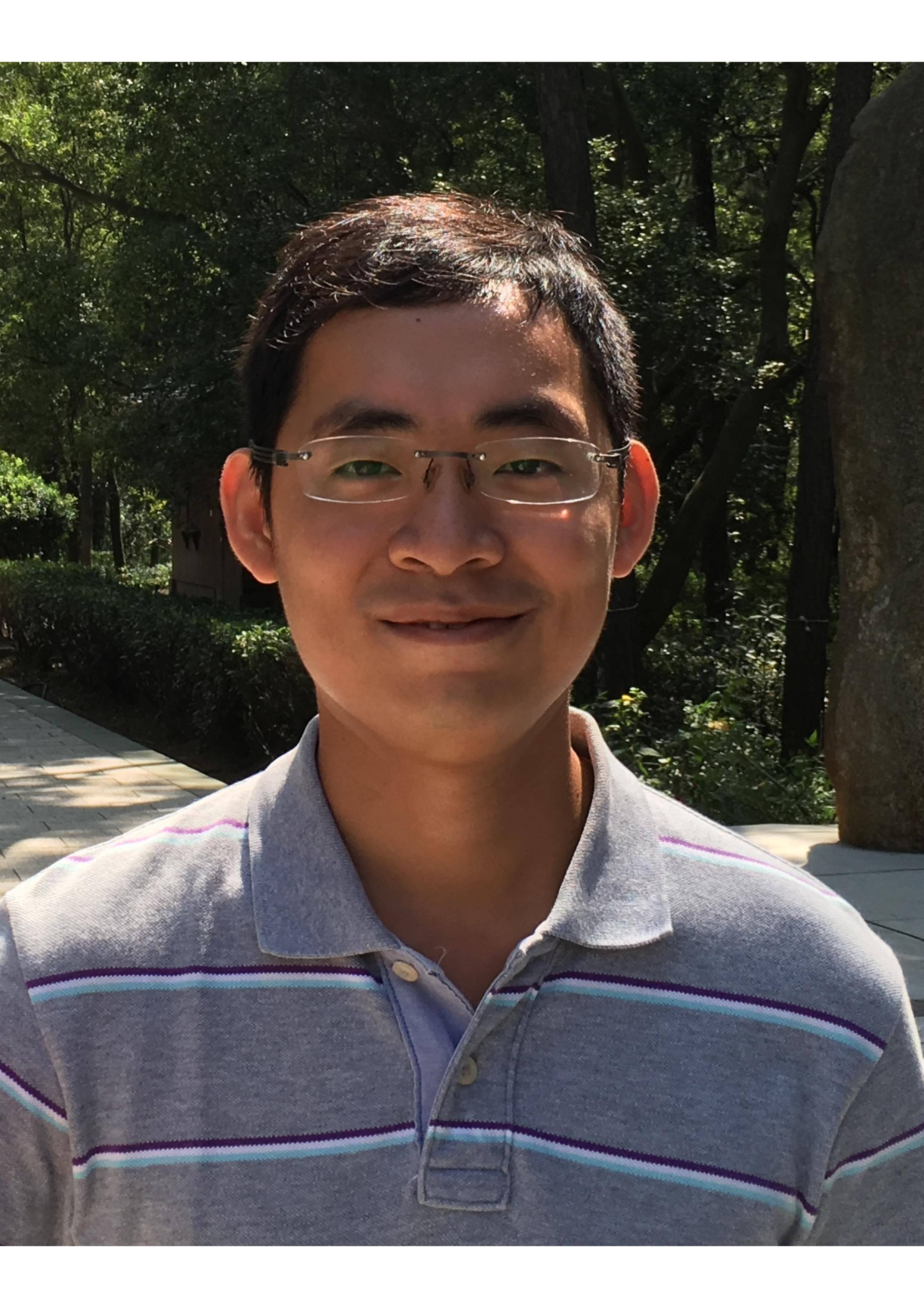}}]{Weng-Tai Su} (Member, IEEE) received the B.S. degree in electrical engineering from the National Yunlin University of Science and Technology, Yunlin, Taiwan, in 2012, Taiwan, the M.S. degree in electrical engineering from National Tsing Hua University (NTHU), Hsinchu, Taiwan, in 2014. He is currently pursuing his Ph.D. degree at the Department of Electrical Engineering of NTHU.
	
His research interests mainly lie in machine learning, image and video processing, and computer vision. 
\end{IEEEbiography}

\vspace{-0.4in}
\begin{IEEEbiography}
	[{\includegraphics[width=1in,height=1.25in,clip,keepaspectratio] {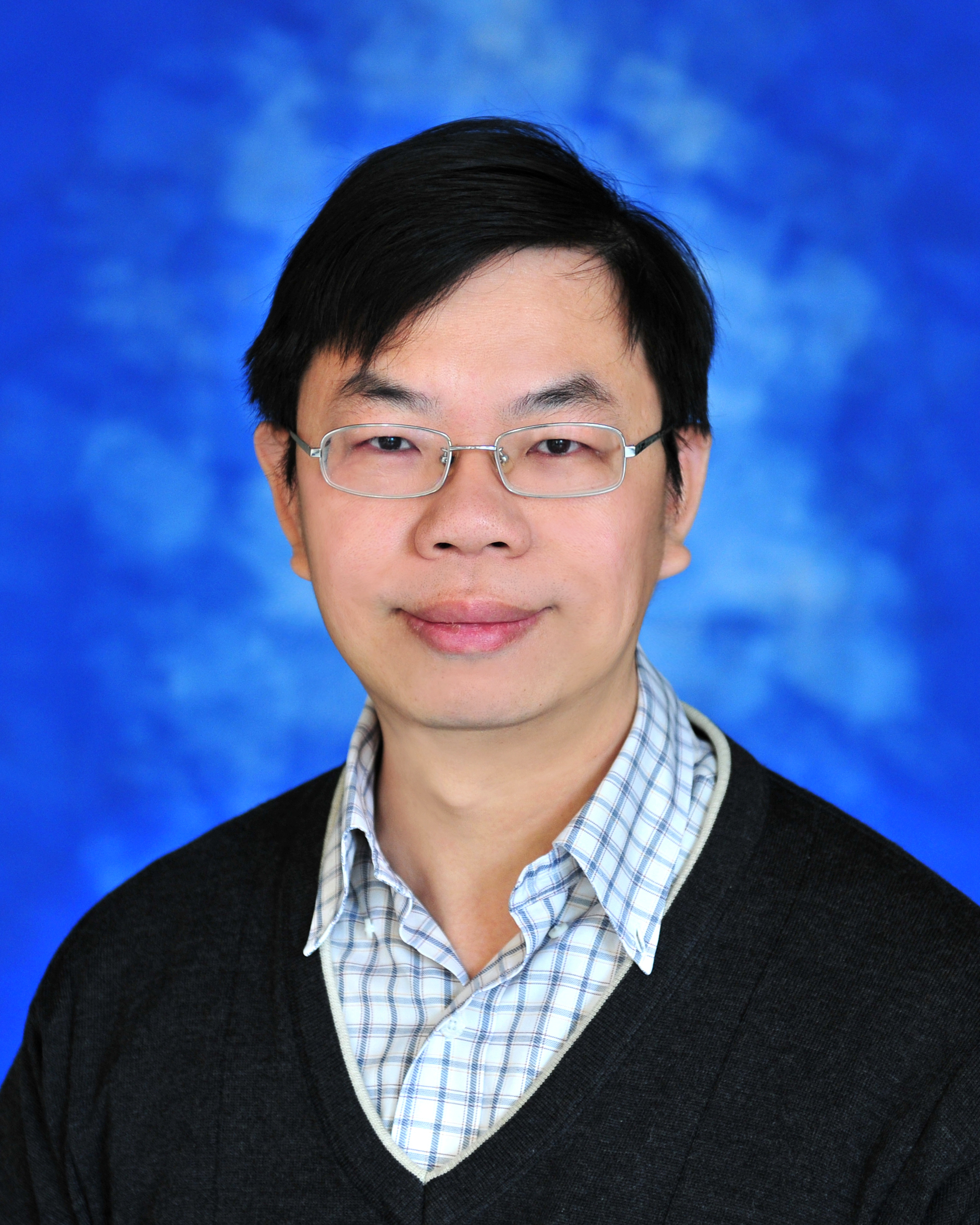}}]
	{Chia-Wen Lin}
	(Fellow, IEEE) received his Ph.D. degree from National Tsing Hua University (NTHU), Taiwan, in 2000.  
	Dr. Lin is currently a Professor with the Department of Electrical Engineering and the Institute of Communications Engineering, NTHU.   His research interests include image/video processing, computer vision, and machine learning.  He has served as Fellow Evaluating Committee member (2021), BoG Member-at-Large (2022--2024), and Distinguished Lecturer (2018--2019) of IEEE Circuits and Systems Society.   He was Chair of IEEE ICME Steering Committee (2020--2021). He served as TPC Co-Chair of IEEE ICIP 2019 and IEEE ICME 2010, and General Co-Chair of IEEE VCIP 2018. He was a recipient of Outstanding Electrical Engineer Professor Award presented by the Chinese Institute of Electrical Engineering, Taiwan. He received two best paper awards from VCIP 2010 and 2015. He has served as an Associate Editor of \textsc{IEEE Transactions on Image Processing}, \textsc{IEEE Transactions on Circuits and Systems for Video Technology}, \textsc{IEEE Transactions on Multimedia}, and \textsc{IEEE Multimedia}.  He served as a Steering Committee member of \textsc{IEEE Transactions on Multimedia} from 2013 to 2015.
\end{IEEEbiography}

\vspace{-0.4in}
\begin{IEEEbiography}
	[{\includegraphics[width=1in,height=1.25in,clip,keepaspectratio] {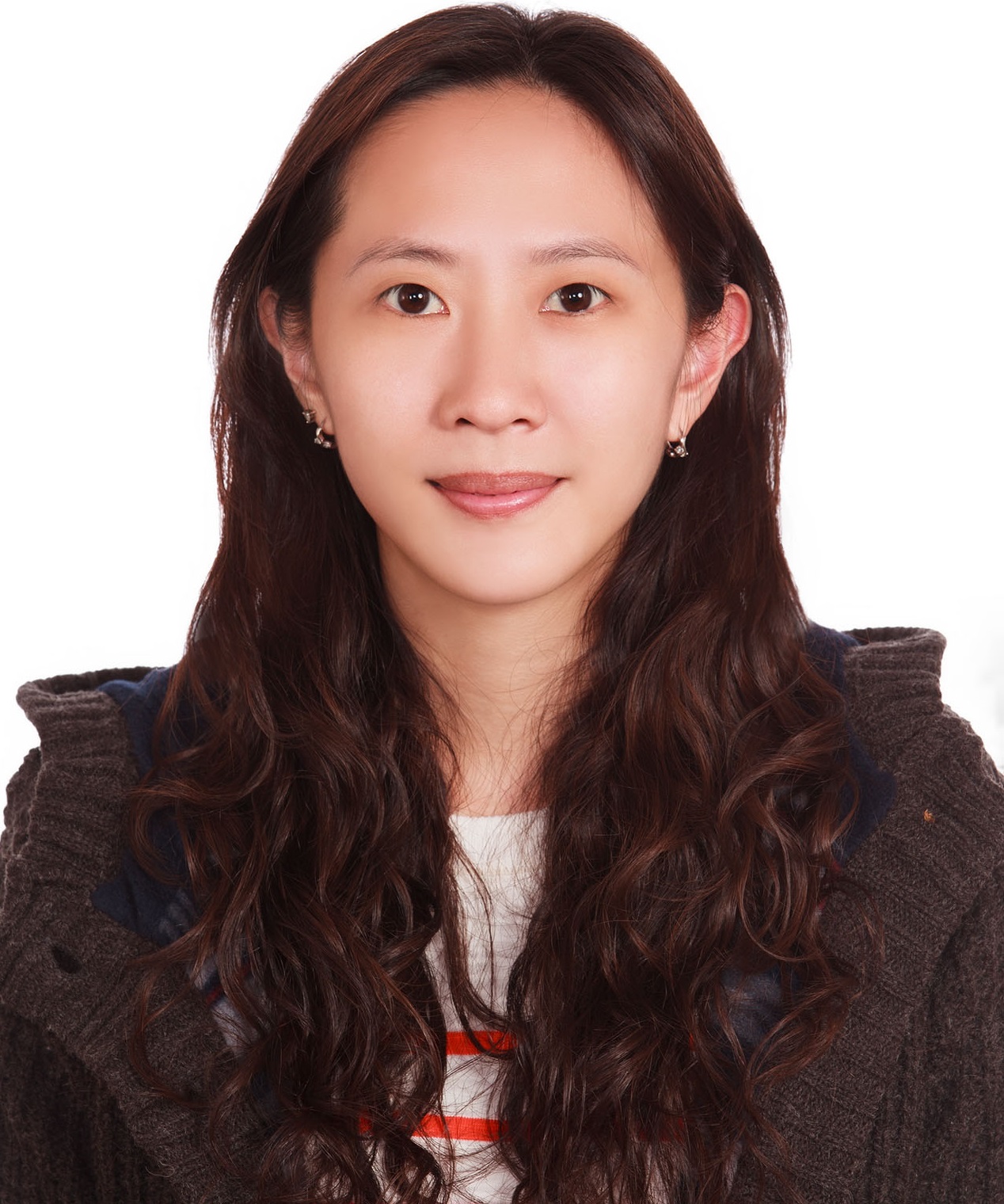}}]
	{Shao-Yun Fang}
	(Member, IEEE) received the B.S. degree in electrical engineering from National Taiwan University (NTU), Taipei, Taiwan, in 2008 and the Ph.D. degree from the Graduate Institute of Electronics Engineering, NTU in 2013. She is currently a Professor of the Department of Electrical Engineering, National Taiwan University of Science and Technology (NTUST), Taipei, Taiwan. Her current research interests focus on physical design and design for manufacturability for integrated circuits. Dr. Fang was the recipient of two Best Paper Awards from the 2016 International Conference on Computer Design and the 2016 International Symposium on VLSI Design, Automation, and Test, and two Best Paper Nominations from the 2012 and 2013 International Symposium on Physical Design.
\end{IEEEbiography}

\vspace{-0.4in}
\begin{IEEEbiography}
	[{\includegraphics[width=1in,height=1.25in,clip,keepaspectratio] {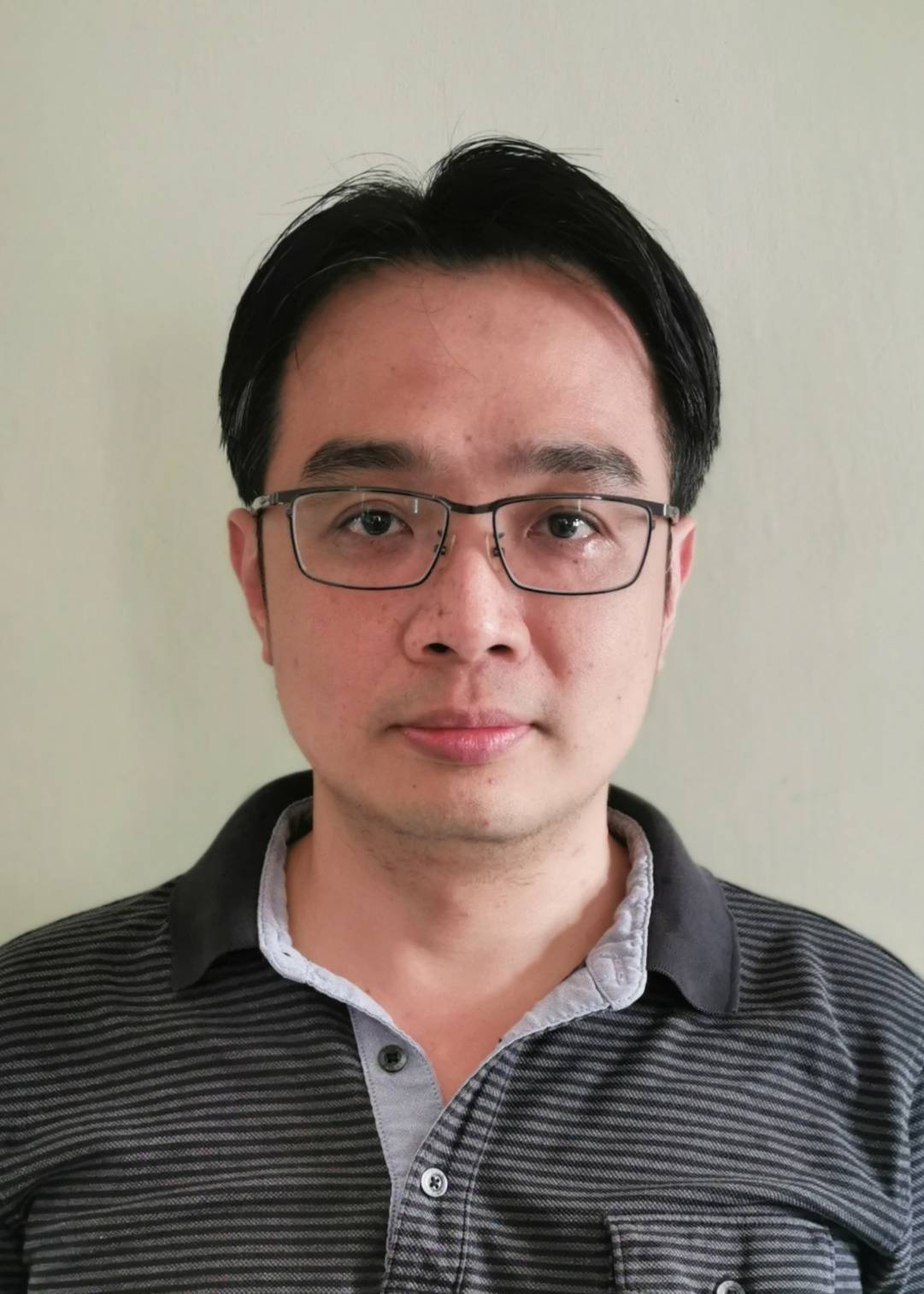}}]
	{Pin-Yian Tsai}
    received his M.S. degree in Physics from National Tsing Hua University (NTHU), Taiwan, in 2008. He is currently a technical manager of the Product Engineering Department in United Microelectronics Corporation (UMC). He led the launch of UMC’s first 14nm product tape out (2017) and is currently working and researching on the field of Design for Manufacturing (DFM). He is now focusing on developing methods for predicting weak patterns in layout manufacturing and automatic optical proximity correction (OPC) to improve the manufacturing yield.
\end{IEEEbiography}

\vspace{-0.4in}
\begin{IEEEbiography}
	[{\includegraphics[width=1in,height=1.25in,clip,keepaspectratio] {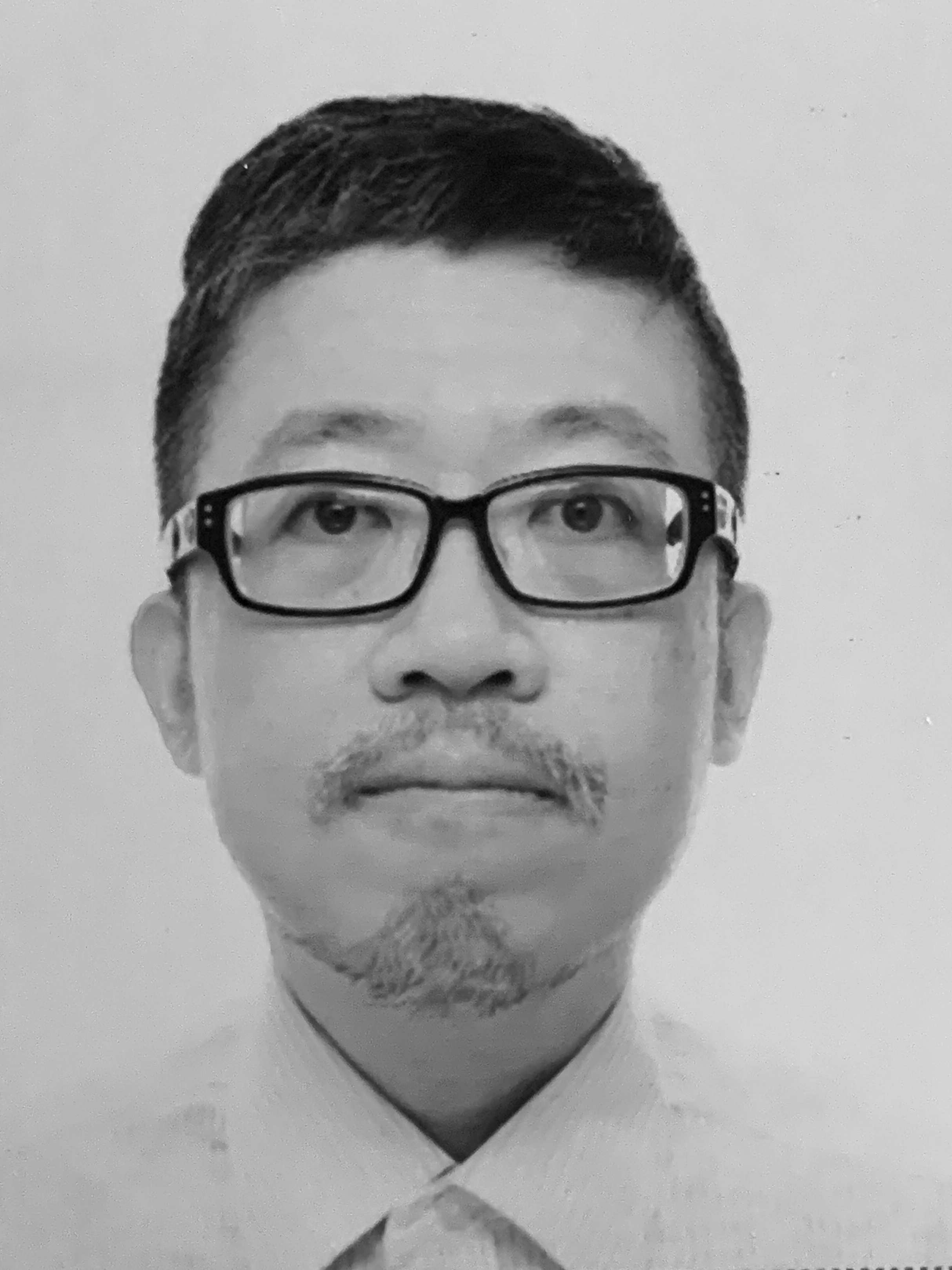}}]
	{Yan-Hsiu Liu}
	received his M.S. degree in Chemistry from National Tsing Hua University (NTHU), Taiwan, in 2002. In 2004, he joined United Microelectronics Corporation (UMC) as a process integration engineer in Hsinchu, Taiwan. He is currently working as a deputy department manager on the development of smart manufacturing and responsible for industry-academia cooperation/collaboration. His research interests include the areas of intelligent manufacturing systems, adaptive parameter estimation, and neural networks.
\end{IEEEbiography}

\end{document}